\PassOptionsToPackage{table}{xcolor}
\documentclass[12pt]{article}

\usepackage{scicite}

\usepackage{times}
\usepackage{tikz}
\usepackage{lscape}

\usepackage{amsmath}

\usepackage{color, colortbl}

\usepackage{threeparttable}
\usepackage{multirow}

\usepackage{bbding}
\usepackage{pifont}
\usepackage{wasysym}
\usepackage{amssymb}

\usepackage[table]{xcolor}
\definecolor{lightgray}{rgb}{0.92, 0.92, 0.92}
\definecolor{green}{rgb}{0.0, 0.62, 0.38}
\definecolor{red}{rgb}{0.84, 0.00, 0.25}

\usepackage{nameref}
\makeatletter
\newcommand*{\currentname}{\@currentlabelname}
\makeatother

\usepackage{comment}
\usepackage{graphicx}
\usepackage{setspace}
\usepackage[font=footnotesize,labelfont=bf,figurename=Fig.,labelsep=period]{caption}

\newcommand{\beginsupplement}{%
        \setcounter{figure}{0}
        \setcounter{table}{0}
        \setcounter{equation}{0}
        \renewcommand{\thefigure}{S\arabic{figure}}%
        \renewcommand{\thetable}{S\arabic{table}}%
        \renewcommand{\theequation}{S\arabic{equation}}%
     }

\newenvironment{sciabstract}{%
\begin{quote} \bf}
{\end{quote}}

\title{Self-deployable contracting-cord metamaterials\\ with tunable mechanical properties} 

\author
{Wenzhong Yan,$^{1,2*}$ Talmage Jones,$^{2}$ Christopher L. Jawetz,$^{2,3}$ Ryan H. Lee,$^{2}$ \\Jonathan B. Hopkins,$^{2}$ and Ankur Mehta$^{1}$ \\
\\
\normalsize{$^{1}$ Electrical and Computer Engineering Department, UCLA, USA.}\\
\normalsize{$^{2}$ Mechanical and Aerospace Engineering Department, UCLA, USA.}\\
\normalsize{$^{3}$Woodruff School of Mechanical Engineering, Georgia Tech, USA.}\\
\normalsize{$^\ast$To whom correspondence should be addressed:}\\
\normalsize{wzyan24@g.ucla.edu au}
}
\date{}

\begin{document} 


\baselineskip24pt


\maketitle

\begin{sciabstract}
Recent advances in active materials and fabrication techniques have enabled the production of cyclically self-deployable metamaterials with an expanded functionality space. However, designing metamaterials that possess continuously tunable mechanical properties after self-deployment remains a challenge, notwithstanding its importance. Inspired by push puppets, we introduce an efficient design strategy to create reversibly self-deployable metamaterials with continuously tunable post-deployment stiffness and damping. Our metamaterial comprises contracting actuators threaded through beads with matching conical concavo–convex interfaces in networked chains.
The slack network conforms to arbitrary shapes, but when actuated, it self-assembles into a preprogrammed configuration with beads gathered together. Further contraction of the actuators can dynamically tune the assembly's mechanical properties through the beads' particle jamming, while maintaining the overall structure with minimal change. We show that, after deployment, such metamaterials exhibit pronounced tunability in bending-dominated configurations: they can become more than 35 times stiffer and change their damping capability by over 50\%. Through systematic analysis, we find that the beads' conical angle can introduce geometric nonlinearity, which has a major effect on the self-deployability and tunability of the metamaterial. Our work provides routes towards reversibly self-deployable, lightweight, and tunable metamaterials, with potential applications in soft robotics, reconfigurable architectures, and space engineering.

\end{sciabstract}

\section*{Summary}
Mechanical metamaterials can reversibly self-deploy and have continuously tunable mechanical properties after deployment.

\section*{Introduction}
\label{sec1}

Self-deployment is widespread in nature, with examples as varied as earwig wings and peacock spider flaps \cite{vincent2001deployable}. Specifically, earwigs' self-deployable wings  allow for both a large-area shape during flight and a compact, folded package when navigating tight underground habitats \cite{faber2018bioinspired}. Given its high energy efficiency, space efficiency, adaptability, and multifunctionality, this transforming strategy is widely seen in art and engineering, with applications spanning architecture, robotics, medical devices, consumer products, and aerospace technologies \cite{pellegrino2001deployable}. The length scales for these applications range from nanometers to meters \cite{whitesides2002self,xu2015assembly,felton2014method,liu2021micrometer,chen2019autonomous,kim2018origami,mallikarachchi2014design}.

Recently, the concept has gained increasing traction in the field of metamaterials \cite{lakes1987foam,zheng2014ultralight,buckmann2015mechanical,kadic20193d,jiao2023mechanical}, propelled by advancements in functional materials and sophisticated fabrication techniques, to achieve material-level self-deployment on demand \cite{xia2022responsive,bertoldi2017flexible}. Typical construction principles for self-deployable metamaterials include the use of linkages \cite{wang2016deployable,you2012motion}, origami/Kirigami inspired methods \cite{meng2023cage}, and tensegrity-based approaches \cite{shah2022tensegrity,spiegel2023shape}. The transformation between configurations is often driven by phase transition \cite{kim2015self},
strain mismatch \cite{meng2023cage}, and mechanical instability \cite{meng2022deployable,fu2018morphable}. These driving mechanisms can be triggered by controllable physical signals, including electric current \cite{hawkes2010programmable,sun2023embedded}, temperature \cite{liu2017programmable,kotikian2019untethered}, magnetic fields \cite{kim2018printing}, and pneumatic pressure \cite{melancon2021multistable}. Once deployed, however, the metamaterials’ mechanical properties are usually fixed, making each metamaterial suitable only for a specific task and limiting its applicability in unpredictable, complex environments \cite{zhai2018origami,zhai2020situ,zappetti2020phase,park2014design,zappetti2020variable}.
Mechanical metamaterials with both variable stiffness and self-deployability have been demonstrated, but the two features in these materials are often coupled \cite{meng2022deployable,overvelde2017rational,kim2018origami}. Consequently, developing mechanical metamaterials that not only can self-deploy but also retain the ability to continuously tune their mechanical properties post-deployment presents a challenge.

\begin{figure*}[t!]
  \centering
\includegraphics[trim=0in 7.2cm 0in 0in, clip=true, width=\textwidth]{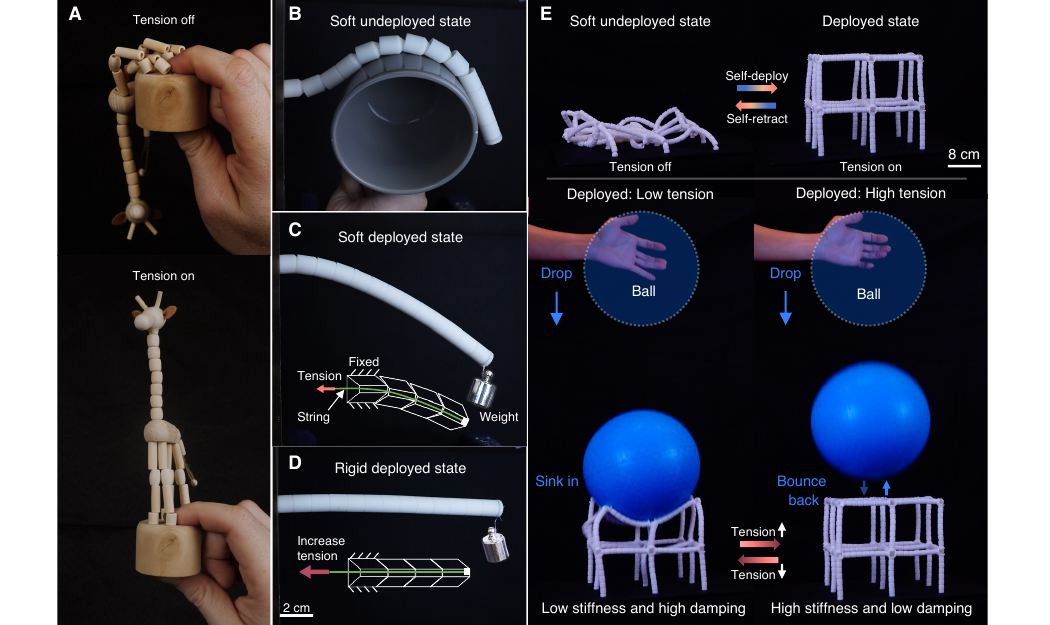}
  \caption{\textbf{The concept and prototype of the self-deployable contracting-cord mechanical metamaterials.} \textbf{(A)} The metamaterial is inspired by push puppets. \textbf{(B-D)} The self-deployment and mechanical properties tuning process of a fundamental beam of the proposed mechanical metamaterials. (B) Image of an undeployed beam---composed of beads threaded by a contracting string-like actuator---in the soft state. (C) Self-assembly of the beads shown in b, after a contracting tension is applied through the actuator. (D) Image of the jammed beam, becoming a stiff load-bearing structure with further contraction of the actuator. \textbf{(E)} Self-deployment and mechanical property tuning of a $2\times2\times2$ cubic lattice. Top: The lattice changes from a soft unassembled state to a deployed state with tension generated by the embedded actuators. Bottom: After deployment, a lattice with low string tension can capture a dropped ball by dissipating the kinetic energy through extensive damping. However, a ball dropped on the same lattice with high string tension will bounce back due to its increased stiffness. Note that the weights attached to the beam in (C) and (D) are selectively brightened in the images for visualization purposes.}
  \label{fig:1}
\end{figure*}

The realization of these self-deployable mechanical metamaterials could allow devices and machines to be stored and transported in retracted, compact states and then self-assembled to the intended configurations in situ. Subsequently, their mechanical properties can dynamically adjust with minimal changes in configuration, enabling them to adapt to various conditions, such as differing vibration frequency and amplitude, surface roughness, or contact stiffness. For example, a self-deployable soft robot, after assembly, can tune its limbs' stiffness to accommodate different terrains while retaining its body structure for optimal locomotion performance \cite{baines2022multi,maccurdy2016printable}. Other potential applications include impact-resistant self-assembling shelters (with enclosed shells) for air-dropping into disaster areas \cite{costa20224d,zhao2023starblocks}, compact vibration insulators with programmable damping in dynamic environments \cite{yan2015experimental,hong2015experimental}, and more.

To achieve such self-deployable mechanical metamaterials, a fundamentally new design paradigm is required. (i) An efficient structural construction principle---using a single actuation system for both self-deployment and mechanical properties tuning---is favorable for minimizing implementation complexity and weight. (ii) Once metamaterials are deployed, it is advantageous to maintain minimal structure variation over the tuning of mechanical properties. This is because large structure (or configuration) change might introduce undesired interference for a given application. (iii) Self-deployment should be reversible and reusable, allowing cycling and long-term operations.  

Here, we present mechanical metamaterials that can reversibly self-deploy and possess post-deployment tunable stiffness and damping based on the proposed contracting-cord particle jamming (CCPJ), inspired by push puppets (Fig. \ref{fig:1}A). The metamaterial consists of networked chains where beads with concavo-convex interfaces are threaded along contracting actuators. \textcolor{black}{These uniquely engineered beads with interlocking interfaces, along with contracting-cord actuators, enable highly precise self-deployability and a broad range of tunability in both stiffness and damping within 3D metamaterials (see Supplementary Table \ref{table:compare}).}
We explore CCPJ-based beams experimentally and numerically by varying applied contracting tension. We also compare the results over the geometric parameter space against the underlying physics of the beads and beam. We show that a self-deployed beam has more evident tunability in its bending-dominated configuration: with a small external contracting tension (120 N), they become more than 35 times stiffer and achieve a 52\% change in damping capability compared with their relaxed configuration. By varying the interfacial conical angle, the beam's self-deployability (including the alignment accuracy and success rate of assembly) and mechanical property tunability vary vastly, due to the nonlinearity arising from geometric and frictional interactions between beads. 

\textcolor{black}{We also characterize the mechanical tunability of CCPJ-based cubic unit cells that composed of identical unit beams, indicating the viability of using our mechanism to construct arbitrary configurations while retaining advantageous attributes.
Furthermore, comparing the mechanical properties tunability of bending-dominated and stretching-dominated cells confirms the preference for bending-dominated structures in our CCPJ-based metamaterials. Specifically, our bending-dominated cubic cell exhibits a stiffness change of approximately 32 times and a 40\% reduction in damping. These results show consistent performance with those observed in unit beams.} In addition, we demonstrate the proposed CCPJ-based metamaterials by integrating actuators (including electrically-driven thermal artificial muscles and motor-driven cables) to enable on-demand, rapid self-deployment/self-retraction and stiffness tuning of larger scale metamaterials. Therefore, this research paves the way for a new class of materials that can self-deploy on-demand and dynamically tune their mechanical properties in situ to adapt to their surroundings, bringing metamaterials closer to practical applications.

\section*{Results}
\subsection*{Design and Mechanism}\label{sec:design}

Figure \ref{fig:1}B shows the fundamental unit of our self-deployable contracting-cord mechanical metamaterial, i.e. the particle-jamming beam. Each particle is a solid cylindrical bead with a central hole. Unlike conventional tendon-driven non-concave particle jamming \cite{jiang2019chain,beatini2013cable}, we use beads with matching conical concavo–convex interfaces (Fig. \ref{SMfig:beamGeometry}). This bead design with matching conical concavo-convex interfaces offers two primary advantages over conventional non-concave designs: (i) It facilitates alignment during self-deployment, and (ii) The surfaces provide geometrical interlocking that enhances frictional contact between adjacent beads, which results in a wide range of mechanical property changes as constraints vary \cite{dyskin2001toughening} (see next section for more analysis). These beads are made of resin, which is manufactured with a high-resolution 3D printer based on low force Stereolithography (Form 3+, Formlabs). This printing method can produce individual beads which are fully dense and isotropic with a smooth surface; it can also fabricate beads in a rapid, programmable manner with an ample design space for arbitrary configurations (see Methods).

To apply tension and confine the beads, string-like actuators are needed, which can thread through the beads and contract/shorten upon activation. These actuators should be capable of providing sufficient contracting stroke and stress to act against resistive torque and force during the processes of both deployment and mechanical properties tuning. We chose to use two types of actuators that satisfy these requirements: motor-driven cables (MDCs) and super-coiled polymer actuators (SCPAs, which function similar to shape memory alloy \cite{haines2014artificial}). With initial slack on the actuator, the beam can freely bend, fold, and conform to curved objects (Fig. \ref{fig:1}B).
When activated, the actuator contracts to pull the beads together, forming a tight assembly (Fig. \ref{fig:1}C). We refer to this process as self-deployment (Supplementary Movie S1). During  self-deployment, the contracting actuator must supply enough tension to overcome the opposing forces and torques caused by frictional contact and gravity. Notably, the rotational symmetry of the cone-shaped interface facilitates bead alignment during this process, whereas a non-concave interface would rely solely on the actuator's tension to align the bead holes. This effect reduces inter-facial friction while greatly simplifying beam design and assembly. With further contraction of the actuator, the assembled beads can serve as load-bearing structures through particle jamming (Fig. \ref{fig:1}D) \cite{liu1998jamming}. Complex architectures can be constructed from these basic linear building units. For example, a $2\times2\times2$ cubic lattice could be created by threading beads along its edge topology (Fig. \ref{fig:1}E). This lattice can self-deploy and allow variable mechanical properties after assembly, enabling distinct interactions with external loads such as a dropped ball (see section \nameref{sec:actuation} for more details).

\subsection*{Tuning Mechanical Properties}
\label{sec:tuning}

\begin{figure*}[t]
  \centering
\includegraphics[trim=0in 6.8cm 0in 0in, clip=true, width=\textwidth]{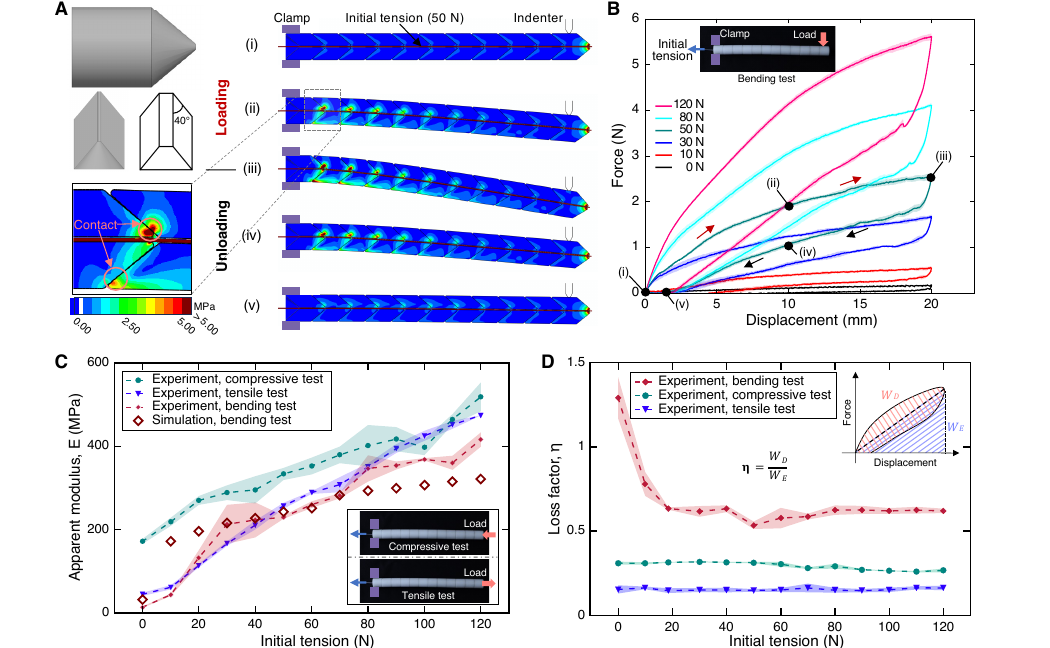}
   \caption{\textbf{Characterization of a CCPJ-based beam with variable applied contracting tension.} \textbf{(A)} The deformation and von Misses stress distribution of a beam under bending loading and unloading. The beam is composed of 11 concavo-convex beads with a 40$^\circ$ cone angle (CAD model shown in the insert) and the string pretensioned at 50 N. \textbf{(B)} Experimentally measured force-displacement curves at different initial contracting tensions. The coloured lines represent the average values, and the shaded areas represent the standard deviations between three different tests. \textbf{(C)} Bending, tensile, and compressive test apparent modulus as a function of the initial contracting tension. The shaded areas represent the standard deviation between three different tests. \textbf{(D)} Loss factor, representing the damping capability, as a function of the contracting tension, for bending, tensile, and compressive tests. The shaded areas are the standard deviation between three different tests.}
  \label{fig:2}
\end{figure*}

To tune the mechanical properties of our assembled beam, we seek to trigger jamming between the beads by applying variable tension at the boundary (Fig. \ref{fig:1}B). One end of the nylon string is fixed to the top bead of the beam. The other end is attached to a force stand. The force stand can adjust the initial tension (small changes in tension occur during testing) applied to the nylon string (see Methods). The applied tension triggers a jamming transition \cite{wang2021structured} and geometrical interlocking \cite{monsef2017mechanical,karuriya2023granular} among beads which increases the frictional and geometrical contact, turning the beams into load-bearing structures. The beam used in these tests has eleven beads with $40^{\circ}$ cone angles (Fig. \ref{fig:2}A, other parameters detailed in Supplementary Table \ref{table:characterization}).

To quantify the change in mechanical properties as a function of contracting initial tension, we perform one-point bending tests and calculate the apparent elastic bending modulus and loss factor (or energy dissipation coefficient) of the beams (see Methods). In these experiments, the sample is clamped at one end, and a line-shaped indenter is applied to the other end in the middle of the top bead (Supplementary Movie 2). 

Before running experiments, we used a finite element (FE) model to visualize the bending process and characterize its underlying mechanism (see Methods). Upon application of a 50 N initial tension on the string, the beam is straight, with stress evenly distributed among the eleven beads (Fig. \ref{fig:2}B(i)). When bent, the stress distribution concentrates more on the beads close to the fixed  end, but remains well-distributed thanks to the additional frictional contact introduced by the interlocking geometry (Fig. \ref{fig:2}A (ii)-(iii)). 
In comparison to the jamming of conventional non-concave beads, our proposed beads with cone-shaped interfaces introduce two contact areas. This delay in separation between beads helps maintain beam stiffness even at large indentation (\ref{SMtext:Comparison40n90deg}). This extended high-stiffness range is advantageous for practical applications where large deformations are often inevitable, and a sudden drop in stiffness could lead to severe failures. During unloading, the tension on the string provides a recovery force to unbend the beam (Fig. \ref{fig:2}A (iii)-(v)). Once fully unloaded, a small residual displacement results from the frictional force between beads (Fig. \ref{fig:2}A(v), see Supplementary Movie S3 for a full animation). The extracted force-displacement curve of the FE beam shows close agreement with the experimental data (Fig. \ref{SMfig:ComparisonModelFEA}); the calculated apparent bending modulus and loss factor also have small deviations (Supplementary Table \ref{table:CmparisonOfModulusandDamping}), suggesting the validity of the FE model. 

We then ran the experiments with various initial tensions on the string. The measured force-displacement curves show initially stiffer regimes at small indentation depths (Fig. \ref{fig:2}B). 
This linear regime is governed by the elastic behaviour of the jammed granular structure \cite{wang2021structured}. As indentation increases, we observe a nonlinear response with a consistently decreasing instantaneous stiffness. This phenomenon is likely due to frictional sliding and local repositioning of the beads \cite{wang2021structured}. Changes in string tension during indentation are small, so their contribution to this nonlinearity can be considered negligible (Fig. \ref{SMfig:tensionVarianceOfString}). 
Although our beams are discrete and strongly anisotropic, we use an apparent elastic bending modulus $E_b$ \cite{strength1979modulus} and loss factor $\eta_b$ \cite{zhang2022dynamic} as parameters to compare the beams’ mechanical properties (see Methods). Specifically, $E_b$ and $\eta_b$ represent the stiffness and the energy dispassion capabilities, respectively. These two parameters can be calculated as:

\begin{equation}
    E_{b} = \frac{16K_b L^3}{3\pi D_{O}^4}
\label{eq:E_b}
\end{equation}
\begin{equation}
  \eta_{b} = \frac{W_D}{W_E}, 	0\leqslant\eta_{b}\leqslant 2
\label{eq:damping_b}
\end{equation}

\smallskip

\noindent where $K_b$ is the stiffness of the initial linear regime from the one-point bending test (Fig. \ref{fig:2}B). $L$ and $D_O$ are the as-fabricated length and outer diameter of the beams (see Methods). $W_D$ is the dissipated/damped energy during the bending process, which is estimated as the area enclosed by the loading curve and the unloading curve. $W_E$ represents the stored energy (see Methods).  

As the initial tension increases from 0 N to 120 N, the apparent bending modulus increases monotonically, from about 12.4 MPa to 434.6 MPa, by over 35 times (Fig. \ref{fig:2}C). Simulations using the FE model were also run and the approximated apparent bending moduli agree relatively well with experimental results. The increase in bending modulus at high tensions is representative of granular materials and is expected since the grains interact by frictional contact \cite{jiang2019chain}.
Under deployed conditions with a small amount of slack present, the stiffness can potentially decrease to an indefinitely low bound; the bending modulus of the sample could be estimated by solely evaluating that of the actuator (i.e., nylon string in this case), which is at the order of $10^{-3}$ MPa and thus makes the stiffness ratio at the order of $10^{5}$. Here we only quantify the stiffness starting from 0 N tension without slack. The loss factor shows a monotonic drop (from 1.29 to 0.62) at the low tension region and quickly reaches a saturation value as the tension continuously increases, approximately a 52\% reduction (Fig. \ref{fig:2}D). This is because the increasing compressive stress between beads enhances their frictional contact and shifts the major deflection mode from high-damping sliding to low-damping elastic behaviour. 
In addition to the wide range of damping variance, the jammed beam shows an overall large loss factor of above 0.6, representing a high damping material  \cite{lihua2019programmable,ma2016tunable}.

To better understand the performance of the beam, we also perform tensile and compressive tests (Supplementary Movie S4 and S5) and calculate the apparent tensile/compressive modulus and loss factors (see Methods). The measured force-displacement curves (Fig. \ref{SMfig:tensileTestOverAngles} and \ref{SMfig:compressiveTestOverAngles}) both show initially linear regimes at small indentation depths and nonlinear responses at large deformation, which is similar to the behavior observed in the bending tests. We obtained increases of about 10 times and 3 times for apparent tensile and compressive modulus, respectively, as the initial tension increases from 0 N to 120 N (Fig. \ref{fig:2}C). Here, these increases mostly originate from the intrinsic non-linearity of the constituting materials (nylon and resin) and geometry. The nylon string material has much larger non-linearity than the resin bead material, giving a higher stiffness tunability for samples under tensile loading (Fig. \ref{SMfig:nylonTest} and \ref{SMfig:resinTest}).
In addition, the damping in both tensile and compressive tests are low and show very limited tunability (Fig. \ref{fig:2}D). Therefore, by not relying on the nonlinearity of constituting materials, our proposed structured beams have much larger tunability of mechanical properties in bending-dominated configurations rather than stretching-dominated ones. This indicates that we should utilize the bending-dominated mode of the beams to improve the tunability of mechanical properties in the construction of large scale metamaterials.

\begin{figure*}[t!]
  \centering
\includegraphics[trim=0in 7cm 0in 0in, clip=true, width=\textwidth]{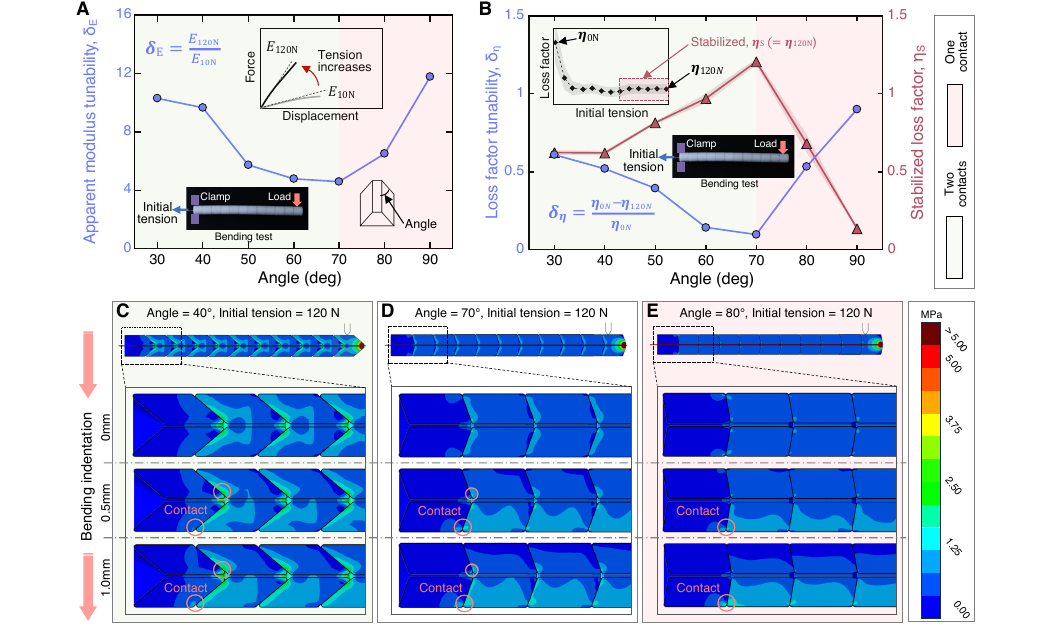}
  \caption{\textbf{Relating the tunability of mechanical properties to cone angle.} \textbf{(A)} The tunability of the apparent bending modulus, $\delta\textsubscript{E}$, as a function of bead cone angle, $\alpha$, as determined by experimental characterization. $\delta\textsubscript{E}$ is defined as the ratio of the apparent bending modulus at 120 N over the modulus at 10 N. \textbf{(B)} The tunability of loss factor, $\delta_{\eta}$, as a function of bead cone angle for bending tests. The shaded areas represent the standard deviation between three different tests. See Methods for a detailed definition of $\delta\textsubscript{E}$ and $\delta_{\eta}$. \textbf{(C-E)} The simulated deformation and von Misses stress distribution of three beams (with beads having 40$^\circ$, 70$^\circ$, and 80$^\circ$ cone angles) under bending indentation up to 1 mm. The strings are pretensioned to 120 N.}
  \label{fig:4}
\end{figure*}

\subsection*{Effect of the Design Parameters}
\label{sec:effect}
The performance of our contracting-cord mechanical metamaterials depends on many factors, including geometry, dimensions, and material properties \cite{jiang2019chain}. We identify the beads' cone angle as the key parameter affecting both self-deployment and mechanical properties tunability (the ability to widely tune the mechanical properties) \cite{karuriya2023granular}. \textcolor{black}{The effects of other factors, including the beads' Young's modulus, edge radius, beads' length, and friction coefficients, are discussed in \ref{SMtext:ParametricFEA} and \ref{SMtext:effectLength}.}

To explore the relationship between the cone angle and the jammed structure’s mechanical properties, we experimentally studied the mechanical responses of beams (with the beads from 30$^\circ$ to 90$^\circ$ with an interval of 10$^\circ$) under one-point bending at different contracting initial tensions (Fig. \ref{SMfig:bendingTestOverAngles}). We also studied the mechanical responses of these beams under tensile and compressive tests, however, these tests did not exhibit as significant of a tunability phenomenon as in the bending tests (Fig. \ref{SMfig:tensileTestOverAngles} and \ref{SMfig:compressiveTestOverAngles}), which confirms the conclusion in the last section. 

For bending tests, we observed that the tunability of the apparent bending modulus and the tunability of the loss factor show similar trends as cone angles are varied (Fig. \ref{fig:4}A and B). When the angle is small (less than 70$^\circ$), the neighboring beads permit more complex interactions (i.e., face–face and face–edge contacts) due to their interlocking geometry (Fig. \ref{fig:4}C). As each bead-to-bead interface contains two distinct contact regions, the beam's tension-controlled stiffness is largely dependent on nonlinear contact effects. The level of nonlinearity increases as the angle decreases, which coincides with the observed phenomenon that the stiffness tunability grows as the angle decreases (in light green in Fig. \ref{fig:4}A). This same phenomenon is often observed in interlocking granular media \cite{athanassiadis2014particle}. There are, however, limitations on the minimum angle value imposed by (i) the feasibility of manufacturing of the beads, and (ii) the excessive contact stresses at the interfaces, which could extensively damage individual beads \cite{karuriya2023granular}. At a 70$^\circ$ angle, the jammed system exhibits a transitional behaviour, where the conditions at the conical interfaces shift from two contact areas to one contact area (Fig. \ref{fig:4}D). When the cone angle is larger than 70$^\circ$, the behaviour of the beam resembles conventional non-concave particle jamming (with one contact), showing no interlocking (Fig. \ref{fig:4}E) and presenting unstable slips at large bending indentation (Fig. \ref{SMfig:bendingTestOverAngles}). 
For the area of interest (30–70$^\circ$), the tunability of damping decreases monotonously as the cone angle increases (in light green in Fig. \ref{fig:4}B). With large cone angles, the beam benefits less from the additional contact and tends to slip under bending indentation, causing large energy dispassion across different tension levels, which is indicated by the high stabilized loss factors. Thus, these beams with large angles have smaller damping tunability.

\begin{figure*}[t!]
  \centering
\includegraphics[trim=0in 8.6cm 0in 0in, clip=true, width=\textwidth]{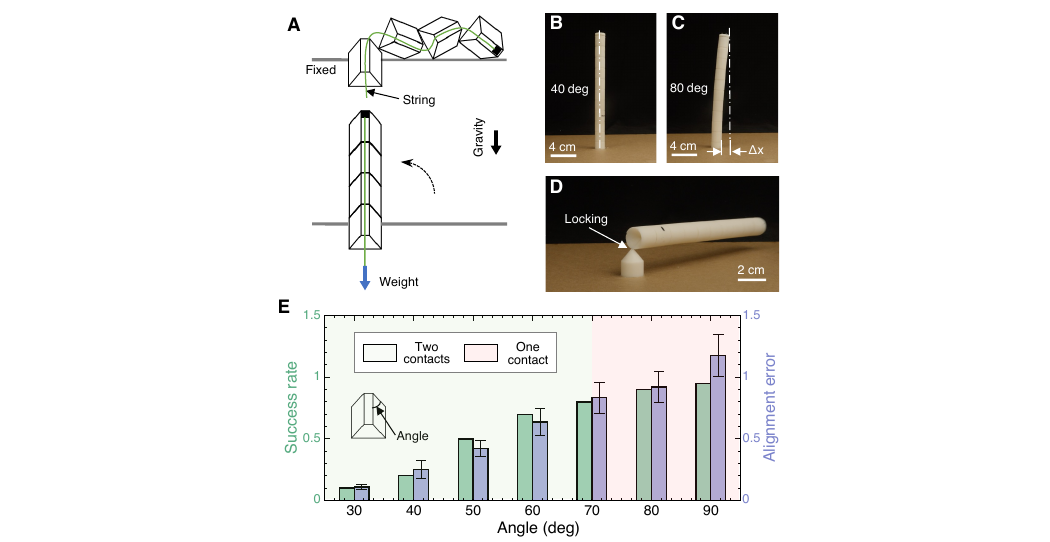}
  \caption{\textbf{Characterizing self-deployment of an individual beam.} \textbf{(A)} Testing schematic. Beads assemble into a vertical beam against gravity, driven by a nylon string under tension induced by dropping a one-kilogram weight. \textbf{(B)} Image of a deployed beam with beads of 40$^{\circ}$ cone angle, showing a high alignment accuracy. \textbf{(C)} Image of an assembled beam with beads of 80$^{\circ}$ cone angle, displaying a large offset, $\Delta x$, between the cap and end beads. \textbf{(D)} A typical deployment failure mode---locking occurs mostly between last two beads.  \textbf{(E)} Success rate and alignment accuracy as a function of cone angle. The success rate is calculated as the ratio of successful attempts over 20 trials. \textcolor{black}{The alignment error is defined as the ratio of the outer diameter of the beads over the offset  ($\Delta x/D_O$)}. Error bars represent the deviation between three different tests.}
  \label{fig:3}
\end{figure*}

Bead cone angle also influences the ease of beam assembly and the alignment accuracy between beads. Here, we vary the bead cone angle (all beads within a single beam have the same cone angle) for self-deployment tests. A one-kilogram weight is tied to the string and released to activate assembly of the beam against gravity (Fig. \ref{fig:3}A, Supplementary Movie S6, see Methods for detailed operation). Figure \ref{fig:3}B and C show successfully assembled beams with 40$^\circ$ and 80$^\circ$ cone angles, respectively. We observed that, at small angles, the last two beads tend to lock easily, which locking cannot be overcome by increasing the tension on the string (Fig. \ref{fig:3}D). This is probably due to the applied string tension having an extremely short moment arm against the end bead, thus failing to overcome the opposing moments from friction and gravity. We use assembly success rate to quantify the ease of self-deployment. \textcolor{black}{
Alignment accuracy is quantified by the alignment error. A small alignment error indicates that the system has high alignment accuracy. The alignment error is defined as the ratio of the misalignment offset to the bead's outer diameter ($\Delta x/D_O$) (Fig. \ref{fig:3}C)}. We found that the assembly success rate monotonously increases as the angle increases from 30$^\circ$ to 90$^\circ$ with an interval of 10$^\circ$. Contrarily, sharper cone angles facilitate bead alignment, which can result in more accurately aligned, and thus more functional, structures (Fig. \ref{fig:3}E). These two opposite trends caused by cone angle indicate a necessary trade-off between alignment accuracy and success rate for certain self-deployment tasks. 

In summary, we explore the complex relationship between beam design parameters and mechanical characteristics. Our results indicate that the bead cone angle controls a trade-off between mechanical properties tunability and self-deployability, which must be considered based on the specific design application.

\begin{figure*}[t!]
  \centering
\includegraphics[trim=0in 8.1cm 0in 0in, clip=true, width=\textwidth]{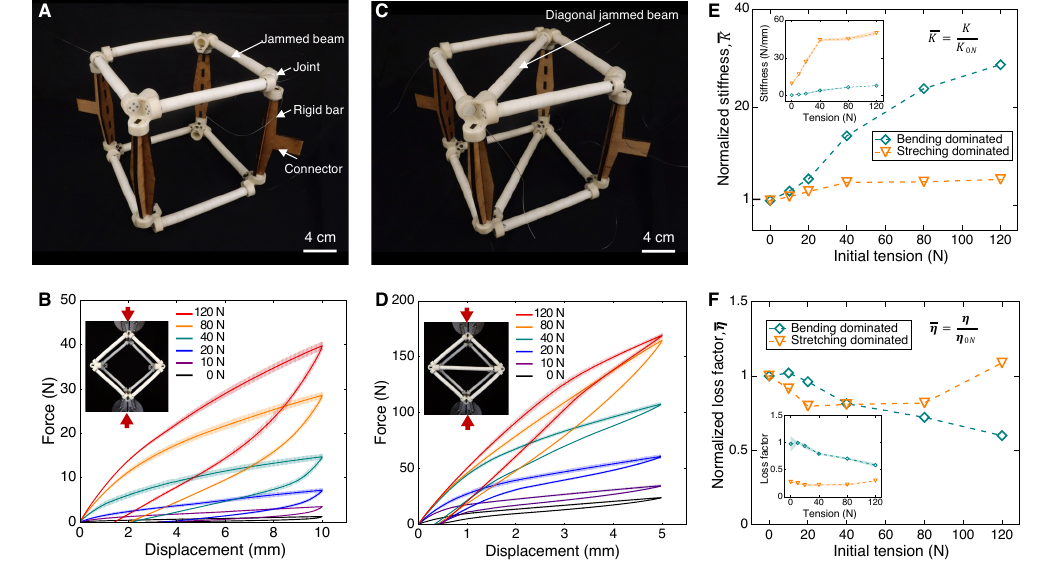}
  \caption{\textbf{Characterization of a single metamaterial unit cell.} \textbf{(A)} Labeled image of a bending-dominated lattice. Each beam in the lattice was assembled with a pre-tensioned nylon string. Beams were connected into two squares using customized 3D printed joints. Four rigid bars were used to connect these two squares. Two of the rigid bars provide connectors for interfacing with Instron clamps. \textbf{(B)} Measured force-displacement curves at different contracting tensions for the bending-dominated lattice. The coloured lines represent the average values, and the shaded areas represent the standard deviation between three different tests. \textbf{(C)} Image of a stretching-dominated lattice. Two diagonal beams distinguish its structure from the bending-dominated lattice. \textbf{(D)} Measured force-displacement curves at different contracting tensions for the stretching-dominated lattice. The coloured lines represent the average values, and the shaded areas represent the standard deviation between three different tests. \textbf{(E)} Normalized stiffness, $\overline{K}$, of lattices as a function of contracting tension. The stiffness at different contracting tensions is normalized over the stiffness at 0 N tension. The insert shows non-normalized stiffness. \textbf{(F)} Normalized loss factor, $\overline{\eta}$, as a function of contracting tension. The loss factor is normalized over the value at 0 N tension. The insert shows the non-normalized loss factor.}
  \label{fig:5}
\end{figure*}

\subsection*{Characterizing Cubic Unit Cells}\label{sec:unitcell}
Our CCPJ-based beams are fundamental building blocks, which can be assembled into lattices for various applications. Here we demonstrate this capability using 40$^\circ$ beads to create two classes of lattices: bending-dominated cubes and stretching-dominated cubes. A bending-dominated lattice consists of eight CCPJ beams, arranged into two squares on opposite sides of a cube (Fig. \ref{fig:5}A). Four rigid bars are used to connect these two squares between corresponding nodes. The stretching-dominated lattice differs from the bending-dominated, in that it has two additional diagonal CCPJ beams (Fig. \ref{fig:5}B). To quantify the characteristics of these lattices, we conducted compressive tests (Supplementary Movie S7) and extracted the force-displacement curves as we varied the contracting tension of each beam (see Methods). For the bending-dominated lattice, the measured force-displacement curves (displacement controlled, in Fig. \ref{fig:5}C) have initially linear regimes at small indentation while nonlinear responses are observed as displacement increases. As tension increases, the stiffness of the bending-dominated lattice shows a large change while that of the stretching-dominated only has a small increase (Fig. \ref{fig:5}D). Specifically, the bending-dominated lattice shows an approximately 32 times increase in stiffness as the initial tension changes from 0 N to 120 N. This variance reflects the large stiffness tunability of a single beam (Fig. \ref{fig:5}E). In contrast, the stretching-dominated lattice exhibits much higher stiffness values \cite{deshpande2001foam} but limited tunability (about 5 times increase) as most beams are in compression. We note that the stiffness of stretching-dominated lattice quickly plateaued at low contracting tensions. 

The bending-dominated lattice shows higher damping capability and tunability than the stretching-dominated one, which is consistent with our conclusion from the preceding section: a single beam has larger damping ability and tunability in bending than it does under tensile/compressive loading (Fig. \ref{fig:5}F). Specifically, the bending-dominated lattice can achieve a $\sim$40\% reduction in its loss factor while the loss factor of the stretching-dominated lattice varies inconsistently and within a smaller range. \textcolor{black}{The tendency of damping of the stretching-dominated lattice is also different from the constructing unit beams since beams are mainly subject to compressive or tensile load, instead of bending load.}

\subsection*{Actuating Functional Metamaterials}
\label{sec:actuation}

\begin{figure*}[ht!]
  \centering
\includegraphics[trim=0in 7.6cm 0in 0in, clip=true, width=\textwidth]{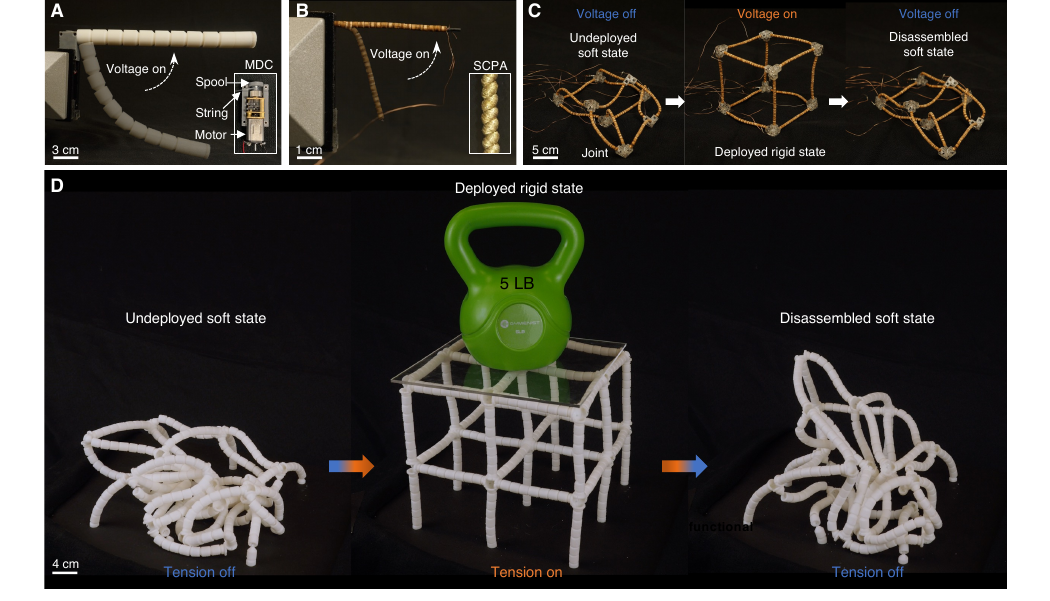}
  \caption{\textbf{Actuation and self-assembly of metamaterials and functional devices.} \textbf{(A)} Self-deployment of a resin-printed beam actuated by an MDC. The insert shows the composition of an MDC--- a DC motor, a spool, and a nylon string. \textbf{(B)} Self-deployment of a plywood beam driven by an SCPA. The insert shows the SCPA, which can be activated electrically or thermally. \textbf{(C)} A cubic lattice can self-deploy and self-collapse by actuating the SCPAs on-demand. The lattice consists of twelve beams (shown in (B)) with eight joints. \textbf{(D)} On-the-fly manipulation of a $2\times2\times2$ cubic metamaterial. The metamaterial is controlled by nylon strings. Once the strings are tensioned, it quickly transforms from its compact, ultrasoft state to its large-volume, load-bearing state, which can sustain a dumbbell of 5 pounds. Upon release, the metamaterial collapses and returns to its original soft state, ready for subsequent operation.}
  \label{fig:6}
\end{figure*}

To autonomously deploy the proposed metamaterials, the actuators need to satisfy several requirements. (i) The contracting parts of the actuators should be string-like to fit into the holes of the beads. (ii) The actuators should be soft and elastic, enabling compact storage, impact resilience, and reversible operation. (iii) The actuators need to be capable of exerting sufficient contracting strain and stress. Here, we choose to use MDCs and SCPAs as example actuators to demonstrate the feasibility of implementing our metamaterial. Each MDC module consists mainly of a motor, a nylon string, and a customized spool (see Methods). When activated, the motor can either pull the string to contract and deploy the beam or release the string to collapse the assembled beam (Fig. \ref{fig:6}A and Supplementary Movie S8). MDCs exhibit excellent power density at the centimeter scale, fast response speed, and an easy integration interface, making them highly suitable for applications at this scale. SCPAs are conductive and can be electrically/thermally driven, which gives rise to a wide range of applications. To avoid overheating, we used flat plywood beads with this actuator. Once electrical current is applied, the soft assembly of beads becomes rigid (Fig. \ref{fig:6}B). Upon cutoff of the applied electricity, the beam softens and drops under gravity (Supplementary Movie 9). Smart material-based actuation holds promise for challenging applications, such as small-scale deployments and remote operations in extreme environments, such as areas with strong magnetic fields. Additionally, we assembled twelve such beams into a cube. The cube demonstrated successful self-deployment and subsequent self-collapse controlled by electrical signal, expanding its occupied volume by about \textcolor{black}{15} times (Fig. \ref{fig:6}C, Supplementary Movie S10). This feature is especially important for remote deployment applications, where materials may need to be transported in a compact volume and assembled autonomously in situ to form functional machines for given tasks \cite{felton2014method}. 

Next, we demonstrated the viability of creating larger-area (and volume) metamaterials by prototyping a cubic lattice composed of an $2\times2\times2$ arrangement of unit cells (Fig. \ref{fig:6}D). We chose to fabricate the lattice with resin beads having 40 degree cone angles for better alignment. For ease of assembly, beads were 3D printed hollow to reduce gravitational forces and their edges were smoothed to avoid locking. For simplicity, we routed several nylon strings through all beads in a specific pattern (see Methods). The soft assembly could be quickly deployed into a large and rigid $2\times2\times2$ lattice. The lattice increased its volume by $\sim$\textcolor{black}{14} times and could sustain a dumbbell of 5 pounds (about \textcolor{black}{13} times its own weight). Upon releasing the tension, the lattice collapsed quickly into its soft state under gravity without requiring external interference (Supplementary Movie S11). 

Finally, we demonstrate the tunability of the same $2\times2\times2$ lattice by comparing its response to an impact load when deployed and under different string tensions (Fig. \ref{fig:1}E, see Supplementary Movie S12 for full process). For each case, we manually drop a ball from a certain height onto the top of the structure. 
When we deploy the lattice and apply a low string tension, the assembly is compliant, allowing the ball to sink into the lattice. The lattice slows the ball to a stop, capturing it and significantly absorbing its kinetic energy. When the tension is increased, the lattice maintains its shape, but increases in stiffness. When dropped, the ball contacts the lattice and bounces back up. After the task, the lattice can self-retract to its disassembled soft, compact state for easy storage and transportation. Notably, we can dynamically and repeatedly shift between the structure's different states (the undeployed state, the deployed compliant state, and the deployed rigid state) by modifying the contracting tension on the actuators.

\section*{Dissusion}
\label{sec:discussion}

We have shown that self-deployable mechanical metamaterials based on contracting-cord particle jamming (CCPJ) can provide on-the-fly continuous tunability of mechanical properties after assembly. \textcolor{black}{These metamaterials are robust to temporary overloading \cite{karuriya2023granular} and resilient to damage \cite{dyskin2001toughening}, attributes that stem largely from their unique configuration. Composed of discrete rigid beads threaded by elastic strings, the system’s compliance allows it to withstand instant overload or collision by dissipating energy through frictional sliding rather than fracturing. Moreover, its discretized structure enables it to sustain the loss of several beads without losing functionality. The proposed metamaterials are easily manufacturable and low cost as well.} In addition, we have systematically explored the underlying mechanics of CCPJ-based beams during both the self-deployment and the jamming transition processes. Notably, these beams features larger tunability in their bending-dominated configuration than in their stretching-dominated mode. The identified key design parameter, bead cone angle, is also investigated, showing a complex effect on mechanical property tunability and self-deployability. This systematic analysis presents the design space and rules for such CCPJ-based metamaterials. 

Deployable lattices with preprogrammed geometry are constructed from concavo-convex beads, demonstrating the viability of creating complex, large-scale CCPJ-based active metamaterials. Recent advances in smart actuators \cite{hong2022magnetically,he2021electrospun} and additive manufacturing make it possible, in principle, to automate the fabrication processes and allow large-scale implementation across various dimensions, targeting different applications. Large-scale active metamaterials hold particular promise for space applications, where the constraints of mass and transportation volume are critical, and reducing the effects of gravity can be advantageous \cite{pellegrino2001deployable,mallikarachchi2014design}. With the integration of power and control, it is possible to envision fully programmable mechanical properties and morphologies via local tuning of each actuator within the metamaterial. This integration could potentially lead to untethered robotic devices for advanced functionalities, such as locomotion, manipulation, and beyond \cite{felton2014method}. In summary, the proposed design paradigm broadens the horizon for designing fully programmable materials, thus offering an impetus to their exploration for practical applications, such as soft robotics, human-machine interaction, medical devices, and space engineering.

\bigskip

\section*{Materials and Methods}
\subsection*{Experiments}
\noindent\textbf{Materials and manufacturing}\\ 
\label{sec:Materials and manufacturing}
The concavo-convex beads are 3D printed via Stereolithography (SLA) 3D printing using a commercial 3D printer (Form 3+, Formlabs). The material used to print the bead is white resin, with a density of 1.15 g cm\textsuperscript{-3}, a tested Young's modulus of 0.571 GPa (\ref{SMtext:CalibratModulus}). 
The contracting cord for experimental characterization is made of nylon string (30LB, Amazon) with a diameter of 0.55 mm. Each string is securely fixed on top beads with screws (Fig. \ref{SMfig:beamAssembly}). Unless otherwise mentioned, testing samples are composed of eleven resin beads with a nylon string. Detailed parameters are shown in Supplementary Table \ref{table:characterization}. To better characterize the behaviors, here we apply tension through a nylon string with a fixed length instead of applying constant tension. This setup more accurately simulates practical scenarios---a string-like actuator with a certain length is employed to generate contracting tension. Note that the tension in the string actuator might vary based on the external loading condition. 
The 90$^{\circ}$ beads are cut using a laser cutter (Speedy 300\textsuperscript{TM} Flexx, Trotec Laser Inc.) from a 3 mm thick sheet of plywood. 

The SCPAs were made using commercially available conductive yarn (235-34 4ply HCB, V Technical Textiles Inc.) with a diameter of about 0.4 mm. These actuators are prepared in two steps (Fig. \ref{SMfig:SCPA}) \cite{yan2022cut}. (i) We insert coils by continuously twisting the conductive yarn under a 280-gram weight. The weight is free to move vertically, but not allowed to rotate. (ii) We anneal the coiled yarn with a cyclic heating/cooling process (0.45 A annealing current, 30 s heating, and 30 s cooling per cycle, 8 hours). The prepared actuators have an average diameter of about 0.71 mm. A single SCPA can generate up to \textcolor{black}{15\%} tensile strain (with pretension) and a maximum force of around \textcolor{black}{3 N}.

The MDC design is adapted from actuators commonly used for tensegrity robots \cite{shah2022tensegrity}. They are primarily composed of a DC motor (1000:1 HP 6V, Pololu), a customized spool, and a nylon sting (Fig. \ref{fig:6} and Fig. \ref{SMfig:MDC}). All three components are housed in a 3D printed case. When the motor runs, the spool on the shaft rotates to shorten the nylon string. Due to the small spool diameter, the MDC can output a maximum tension of about 140 N while the string breaks at about 130 N. This output tension is sufficient for most applications. 

For cubic unit cells (both bending- and stretching-dominated), the end beads were redesigned into two halves (Fig. \ref{SMfig:beamFabLattice}). After the strings were stretched to the desired tension, the other end was clamped by the two halves of the end bead. Screws are used to retain the tension. Then, eight beams with the same applied tension were assembled together with four rigid bars through 3D printed joints (Fig. \ref{SMfig:unitAssembly}). 

For the $2\times2\times2$ cubic lattice, all beads are hollow to reduce the effects of gravity. The bottom edges of the beads are smoothed to improve the ease of self-deployment (Fig. \ref{SMfig:materialsLattice} and \ref{SMfig:assemblyLattice}). To make the structure symmetric and assembly easier, we introduced center beads with both bases concave. We also designed hollow joints that allow nylon strings to pass through with low friction. Thirteen nylon strings in total are routed to go through every single bead in a special pattern (Fig. \ref{SMfig:stringPath}).

\bigskip

\noindent\textbf{Quasi-static mechanical tests}\\
\label{sec:Quasi-static mechanical tests}
The bending characteristics of the beams at different contracting tensions are characterized via one-point bending tests (Fig. \ref{SMfig:bendingTestrig}). Test rig design is described in detail in \ref{SMtext:testRig}. First, the end bead of each beam is clamped in a vise. A 380 mm-long nylon string is then fixed on the load cell of a customized force stand. The load cell is allowed to move to apply a certain tension to the string and then is fastened to the force stand. Different tensions are applied by adjusting the position of force stand's load cell and then fixing it for testing. Note that we straighten the beams before applying an initial tension as gravity can cause a slight bend in the beams before testing. The tests are performed using a universal testing machine (5966, Instron Inc.), with displacement controlled at a loading rate of 10 mm min\textsuperscript{-1}. Three separate tests are repeated at each contracting tension. Before each test, the beams are manually reset to a straight initial configuration. The coloured lines and dashed areas represent the average values and standard deviations for three different tests (Fig. \ref{fig:2}). The deviation observed between the results from different tests at the same tension arises from the initial configurations of the beams, which have different random initial contacts between beads. 

Tensile and compressive tests utilize setups akin to those for bending tests, with the primary difference being the fixtures used (Fig. \ref{SMfig:TandCTest}). Specifically, we rearrange the orientation of the beams and force stand due to the limited space within the testing machine. We place a low-friction pulley within the end bead clamp to reorient the direction of the nylon string's tension. For tensile tests, we designed a connector to grab the top beads and a base to clamp the end beads. For compressive tests, we redesigned the base to allow for direct contact between end beads and the steel clamp of the testing machine, thus eliminating undesired testing errors from fixtures.  

The compressive tests for cubic lattices were conducted in the same machine with their connectors (on rigid bars) clamped onto the grippers of the Instron machine. The tests were run with displacement controlled at a loading rate of 10 mm min\textsuperscript{-1}. Similarly, each tension was repeated three times with the lattices reset to the original configuration between tests. 

\bigskip
\noindent\textbf{Calculating the apparent modulus of CCPJ beams}\\
The stiffness of the initial elastic region in our bending measurement was calculated by fitting the force-displacement curve linearly (Fig. \ref{fig:2}) for small indentation depths (between contact and 0.5 mm). These shallow indentations result in in-plane strains of less than 0.05\%, guaranteeing that the beams experience deformation within their elastic threshold. The apparent bending modulus is computed according to equation (\ref{eq:E_b}), using the measured dimensions and stiffness (slope) of the elastic regime. For apparent tensile and compressive modulus, we apply similar methods according to below equation:
   
\begin{equation}
    E_{t} = \frac{4K_t L}{\pi D_{o}^2}
\label{eq:E_t}
\end{equation}

\begin{equation}
    E_{c} = \frac{4K_c L}{\pi D_{o}^2}
\label{eq:E_c}
\end{equation}

\noindent where $K_t$ and $K_c$ are the respective stiffnesses of the linear regime from the tensile and compressive one-point bending tests (Fig. \ref{fig:2}C and D).

\bigskip
\noindent\textbf{Calculating the stiffness of cubic lattices}\\
We linearly fit the selected regime (ranging from 0 to 1.0 mm) of the force-displacement curves obtained from the compressive measurements (Fig. \ref{fig:5}). The slopes are the stiffness of interest, normalized by the stiffness when the internal tensions of beads are 0 N.

\bigskip
\noindent\textbf{Calculating the loss factor}\\
We programmed a code in Python to integrate the total enclosed areas that represent dissipated energy $W_D$ and the stored energy $W_E$ (Fig. \ref{SMfig:lossFactor}). Specifically, the stored energy is approximated as the sum of half of $W_D$ and the area under the unloading curve according to Ref. \cite{zhang2022dynamic}. The loss factors for all three different tests (i.e., bending, tensile, and compressive) were calculated using equation (\ref{eq:damping_b}). Although there are different ways to define the loss factor, they all yield similar results. Consequently, we choose to only focus on the method stated above.

\bigskip
\noindent\textbf{Calculating the tunability of apparent modulus and loss factor}\\
Tunability refers to the extent to which mechanical properties can be altered in response to increases in the initial tension applied to the contracting cord. For the tunability of apparent modulus, we used the modulus value of the beam at 10 N as the reference. This is because the apparent moduli at low tensions (close to 0 N) are rather unstable. The maximum initial tension on the string is 120 N. Thus the apparent modulus tunability, $\delta_{E}$ is defined as:
   
\begin{equation}
    \delta_{E} = \frac{E_{120N}}{E_{10N}}
\label{eq:tunabilityOfModulus}
\end{equation}

Here, we used the values of loss factors at 0 N as the reference since loss factors are relatively stable even at low tensions. Therefore, the loss factor tunability, $\delta_{\eta}$, is:
   
\begin{equation}
    \delta_{\eta} = \frac{{\eta}_{0N} - {\eta}_{120N}}{{\eta}_{0N}}
\label{eq:tunabilityOfLossFactor}
\end{equation}

\bigskip
\noindent\textbf{Self-assembly test and characterization}\\
We used the same beams (with eleven beads) to characterize the performance of self-assembly. The beams were oriented vertically with the end beads fixed to a rigid platform. A one-kilogram weight was fixed to the end of the nylon string. When tested, the weight was released to generate tension to drive the assembly. We repeated this process 20 times for each angle and calculated the success rate.

The alignment \textcolor{black}{error} is defined according to the equation below:
   
\begin{equation}
    A_{error} = \frac{\Delta x}{D_{O}}
\label{eq:aligAccuracy}
\end{equation}

\noindent $\Delta x$ is the offset between the top bead and end bead (Fig. \ref{fig:3}C), which is extracted from video using the tracking software Tracker (version 5.0.5). $D_O$ is the outer diameter of the beads. The measurement for offset was repeated three times for each angle. Before each test, the beads were randomly shuffled.

\subsection*{Numerical simulations}
\noindent\textbf{Finite element CCPJ beam construction}\\
Finite element (FE) models were created using the commercially available ABAQUS CAE software. The models match the geometry of the CCPJ beams that were experimentally tested on the universal testing machine, where beams are constructed of eleven beads having a cone angle in the range of $30^{\circ}$ to $90^{\circ}$ (Fig. \ref{SMfig:feaWholeAssem}). 

Beads are meshed using 8-node 3D deformable linear brick elements with reduced integration (C3D8R). Each is assigned a linear elastic, isotropic material model with a mass density of 1.15 g cm\textsuperscript{-3} and 0.57 GPa Young's modulus to match the experimentally determined values (\ref{SMtext:CalibratModulus}). A Poisson ratio of 0.4 is also assigned \cite{bodaghi2020reversible}. The string is modeled using a fine mesh of 2-node linear 3D truss elements (T3D2). Its material is modeled as hyperelastic using an Ogden model fitted to experimental data (\ref{SMtext:Ogden}). The indenter is modeled as a 3D analytical rigid body (Fig. \ref{SMfig:feaWholeAssem}).

Strain free adjustments are allowed between beads in the first load step to initiate contact. All contact interactions are assigned a hard normal behavior and a tangential friction coefficient. 
Bead-bead interactions are assigned a coefficient of 0.15 as determined by parametric FE studies (\ref{SMtext:ParametricFEA}). Bead-string and indenter-bead contact interactions are assigned a coefficient of 0.1, which is decided based on the same parametric FE studies (\ref{SMtext:ParametricFEA}). 

\bigskip
\noindent\textbf{Quasi-static bending simulation}\\
Loading conditions in the FE analyses are equivalent to those of the quasi-static bending tests performed on the universal testing machine. The results enable analysis of the underlying mechanics of the experimental response. Each simulation is composed of three load steps: tensioning, indentation, and return. Each step is a quasi-static dynamic implicit procedure with geometric nonlinearity.

During the first step, a tensioning displacement is gradually applied to one end of the string of 380 mm, while the outer surface of the rear bead is fixed in all degrees of freedom (Fig. \ref{SMfig:feaLoadingConditions}). The displacement load corresponds to an applied string tension between 1 N and 120 N  as determined in a separately conducted analysis (Fig. \ref{SMfig:feaPretensionPlot}). The other end of the string is rigidly secured to the tip of the front bead via multi-point constraints. During the deflection step, the indenter is displaced in the negative vertical direction at 10mm/min for 20 mm. During the return step, the indenter returns to zero displacement at the same rate (Supplementary Movie S3 and S13).

\bigskip
\noindent\textbf{Parametric studies}\\
Using custom MATLAB and Python scripts to interface with the ABAQUS FE model, we studied trends in quasi-static bending behavior when varying the following parameters: bead cone angle, string tension, bead Young’s modulus, bead-bead friction coefficient, and bead edge radius (see \ref{SMtext:ParametricFEA} for detailed exploration).

\bigskip

\bibliographystyle{Science}
\bibliography{scibib}

\section*{Acknowledgements}
Our thanks go to Mr. Z. Zheng, Mr. D. Martinez, Dr. R. Lin, and Mr. W. Fernando for their valuable assistance in data analysis and testing.

\clearpage

\cleardoublepage
\pagestyle{empty}

\section*{Supplementary Material}
\maketitle
\title{\textbf{Self-deployable contracting-cord metamaterials with tunable mechanical properties}}

\begin{center}
    
\author
{Wenzhong Yan,$^{1,2*}$ Talmage Jones,$^{2}$ Christopher L. Jawetz,$^{2,3}$ Ryan H. Lee,$^{2}$ \\Jonathan B. Hopkins,$^{2}$ and Ankur Mehta$^{1}$ 

\bigskip

\normalsize{$^{1}$ Electrical and Computer Engineering Department, UCLA, USA.}\\
\normalsize{$^{2}$ Mechanical and Aerospace Engineering Department, UCLA, USA.}\\
\normalsize{$^{3}$Woodruff School of Mechanical Engineering, Georgia Tech, USA.}\\
\normalsize{$^\ast$To whom correspondence should be addressed:}\\
\normalsize{wzyan24@g.ucla.edu}
}
\end{center}

\beginsupplement

\subsection{Comparison of the mechanical behavior of beams with the proposed conical interface versus the conventional non-concave interface}
\label{SMtext:Comparison40n90deg}

\noindent\textbf{Bead contact behavior observations}\\
Here, we use beams with 90$^\circ$ beads to represent the conventional tendon-driven non-concave jamming. We use beams with 40$^\circ$ beads to represent the proposed beads with conical interfaces. The beams studied are subjected to 50 N of initially applied string tension, and have beads with a Young's modulus of 0.571 GPa, 0.05 mm edge radius, a bead-bead friction coefficient of 0.15, and other contact friction coefficient of 0.1. As Fig. \ref{SMfig:40deg-90deg}a shows, the proposed concavo-convex beam has two contact areas at all bead-bead interfaces during bending. These two contact areas can improve the rigidity of the beam even under large indentation (Fig. \ref{SMfig:40deg-90deg}b and Supplementary Movies S3 and S13). This is achieved via geometric interlocking, which causes the stress to be distributed throughout the beam. 

In contrast, the non-concave beads have only one contact area between neighboring beads and the beam's deformation is mainly pivoting around the bottom edge of the end bead (see Supplementary Movie S14 and S15). Consequently, the stiffness decreases dramatically after a certain amount of bending indentation. Such variations in stiffness could lead to significant challenges in practical applications. In addition, the stress concentration on the bottom edge of beads might result in more extensive damage to individual beads. These one-point contact areas are also intrinsically unstable to environmental disturbances, including vibration and impact, which might cause undesired slips and thus large rigidity drop \cite{karuriya2023granular}.

\bigskip
\noindent\textbf{Insight from finite element studies}\\
\noindent\textbf{Observation of contact areas and stress distributions in stress plot animations.}
Figures \ref{SMfig:fea40Bending} and \ref{SMfig:fea90Bending} use von Mises stress plots to illustrate the stability advantage provided by a concavo-convex CCPJ beam over a traditional CCPJ beam. Full animations of the pretensioning, 20 mm deflection, and return cycle can be seen in Supplementary Movies S13 and S14. Supplementary Movies S3 and S15 are animations of the cross-section view shown in the figures. The cross-section view cuts through the center of the beam down its length. The high stress concentration at the tip is a result of the string being secured via beam multi-point constraints (MPCs) at that location. In both models, stresses are highest at the bottom side of the bead closest to the fixed end bead, and then the stresses slowly decrease along the length of the beam. On the $40^{\circ}$ beam, high stresses are also present where the tip of the bead contacts the concave conical surface of the bead in front of it. These stresses also decrease along the length of the bead. The presence of these stresses indicates a higher presence of material deformation in the beads, both due to bending of the tip area and tangential frictional forces. The overall behavior of the $40^{\circ}$ beam is characterized by a more distributed curvature, while that of the $90^{\circ}$ beam resembles more of a hinge with rigid body rotation of the last 10 beads against the fixed end bead. The $40^{\circ}$ beam exhibits sliding along the bead tip surface and distributed stresses in the material, while the $90^{\circ}$ beam exhibits very little sliding and only a localized compression where the largely rigid body rotation occurs. 

\bigskip
\noindent\textbf{Quantification of friction effects in energy plots.}
We confirmed the observed mechanics by plotting the transfer of energy within the system over the full 20 mm range of deflection and return. Fig. \ref{SMfig:feaEnergyPlots}a shows the energy dissipated due to friction in the whole system for 40$^\circ$ beam, along with the internal energy stored due to strain in all beads and in the string. The result for 90$^\circ$ beam is in Fig. \ref{SMfig:feaEnergyPlots}b.

The plots clearly show that the difference in mechanical behavior between the proposed beam and traditional beams is strongly influenced by friction. We see about 0.013 J of energy consumed by friction for a $40^{\circ}$ bead, while a $90^{\circ}$ bead loses around 0.001 J due to friction. The energy stored due to elastic strain in the beads is also about double in the $40^{\circ}$ beam. Also notable is how energy in the beads returns to 0 for a $90^{\circ}$ beam, but in the $40^{\circ}$ case, a small amount remains. This effect can be attributed to the frictional forces equilibrating with the string tension, which attempts to pull the beam back to its original position. As a result, the $40^{\circ}$ bead hysteresis curves return to a 0 N reaction force before the indenter fully returns to its original position. This agrees with the experiment results (Supplementary Movie S2) and animations, which show that the beam does not completely return to its original state and a small amount of compressive stress remains in the concavo-convex beads. The traditional $90^{\circ}$ beams do not exhibit this behavior, and thus the energy stored in the beads returns to 0 J (Supplementary Movie S16).

\bigskip
\noindent\textbf{Discussion of finite element analysis.}
The presence of residual stresses and strain energy in the proposed beads indicates that they may make it harder to straighten out the beam before subsequent bending cycles. The friction maintains strain in the beads, preventing the beam from returning to its complete original state. Note that this effect is distinct from the alignment accuracy in Fig. \ref{fig:4}, which refers to alignment upon deploying the beam (for which angled beads provide an advantage in facilitating alignment). However, the largely consistent stiffness response for the concavo-convex beams provides stability (Fig. \ref{SMfig:fea40Curves}). A traditional $90^{\circ}$ beam, has a very stiff initial response and then dramatically shifts to a low stiffness, while the angled beads create a more prolonged initial stiffness with a more damped return. 

\subsection{Parametric finite element studies}
\label{SMtext:ParametricFEA}

Using finite element (FE) models, we studied the underlying effects of various parameters that are difficult to study experimentally. The results discussed here show trends caused by changes in the following parameters: friction coefficients, bead edge radius, and the Young's modulus of the bead material. Different trends were observed in concavo-convex beads with a cone angle versus non-concave beads with a traditional flat surface are also shown. Here, we study CCPJ beams deflected to 20 mm with $40^{\circ}$ cone angle beads and $90^{\circ}$ cone angle beads.

\bigskip
\noindent\textbf{Determination of friction coefficients at contact interactions}\\
The coefficients of friction (CoF) are difficult to characterize experimentally due to the complex geometry and interaction. Through parametric studies of our FE model, we were able to determine appropriate values. This process also serves to demonstrate the influence of CoF on the mechanical response of the beams.

In the quasistatic bending model, there are three contact interactions of interest: indenter-bead, string-bead, and bead-bead. We assigned each a friction coefficient for tangential motion and studied the effects of varying that coefficient. String tension, bead edge radius, and Young's modulus were kept consistent through all iterations of the friction study with values of 50 N, 0.05 mm and 0.571 GPa, respectively.

Variation of the bead-bead friction coefficient resulted in moderate changes in stiffness and hysteresis \cite{monsef2017mechanical}. We varied the coefficient from 0.1 to 0.2 at an interval of 0.05 and extracted the force-displacement curves (Fig. \ref{SMfig:fea40Curves}a). After calculating the apparent bending modulus and loss factor corresponding to each case, we determined that the curve corresponding to a value of 0.15 is the best fit for the experimental data (Supplementary Table \ref{table:CmparisonOfModulusandDamping}). Unless otherwise stated, this value is assumed to be the default. We also observed that as the bead-bead friction coefficient increases, the stiffness and the loss factor of the beam both increase. Additionally, residual deflection and stresses in the fully returned beam increase. This agrees with the sliding observed in the stress plot animations (Fig. \ref{SMfig:fea40Bending} and Supplemental Movie S3). When sliding occurs, an increased friction coefficient corresponds to an increased stiffness and a larger amount of energy loss in the hysteresis loop.

In the cases of the indenter-bead and string-bead contact interactions, varying the friction coefficient between 0.1 and 0.2 had a negligible effect on the force-displacement curve. However, completely removing these frictional effects caused instability in the FE model. We therefore assigned a value of 0.1 to dampen excessive rigid body movement at these locations.

For beams with $90^{\circ}$ cone angles, varying the bead-bead friction coefficient over the same range also had a negligible effect (Fig. \ref{SMfig:fea90Curves}a). This validates the idea that the bending mechanics of such beams relies less on frictional contact and more on a pivoting motion stiffened by the string.

\bigskip
\noindent\textbf{Effect of bead edge radius}\\
Based on our analysis, we found that bead edges have a moderate impact on beam behavior. As shown in Fig. \ref{SMfig:beamGeometry}, the radius is defined as $R_E$. We empirically determined a minimum radius of $R_E$ to be 0.5 mm: a contact with sharper corner can cause instability in FE models. We varied this radius between 0.5 mm and 1.2 mm in $40^{\circ}$ beams and extracted force-displacement curves (Fig. \ref{SMfig:fea40Curves}b). As the radius decreases, the overall stiffness of the CCPJ beam increases significantly, approaching the experimental behavior of beams with sharp-edged beads. This stiffening phenomenon is commonly observed in particle jamming \cite{karuriya2023granular}. Energy loss within the hysteresis loop appears to be largely unaltered.

The same effect of decreasing radius causing an increased overall stiffness is observed in $90^{\circ}$ cone angle beams (Fig. \ref{SMfig:fea90Curves}b), though the effect is far less pronounced as it is in the $40^{\circ}$ beams.

\bigskip
\noindent\textbf{Effect of bead's Young's modulus}\\
We discovered that the Young's modulus of beads significantly affects the overall beam behavior under bending tests. We varied the modulus between 0.571 GPa and 100 GPa in a $40^{\circ}$ beam and extracted force-displacement curves (Fig. \ref{SMfig:fea40Curves}c). As the modulus increases, the overall stiffness of the CCPJ beam increases. At high modulus values, the beam appears to approach an upper limit of stiffness, converging on a behavior where an extremely high initial stiffness is followed by a sudden transition to a low stiffness. It follows that a model with perfectly rigid beads would exhibit behavior similar to the high modulus cases, featuring dramatic beam stiffness changes and clear separation between beads. Therefore, the Young's modulus of beads should be considered in the model to accurately capture the behaviors of beams.

\bigskip
\noindent\textbf{Discussion on model accuracy}\\
In the force-displacement curves, we see that the FE models agree fairly well with the experimental data obtained on the universal testing machine. Experimental data in each plot is the average of our quasistatic bending test data for 50 N of initially applied string tension. 
Decreasing the radius of the bead edges especially lines the data up well with the experimental data. It is difficult to run a model with sharper edges than 0.5 mm as an extremely fine local mesh is required for computational stability, which comes at a high cost in terms of computational efficiency. 
The lack of hysteresis in the $90^{\circ}$ bead model is likely due to the material model for the string not including hysteresis effects. This is consistent with what the energy plots indicate; the $90^{\circ}$ beams' energy plots show that these beams are especially reliant on the string’s elasticity to hold the structure together (Fig. \ref{SMfig:feaEnergyPlots}b). Friction is the principal source of hysteresis in the $40^{\circ}$ beams, and this was included in the models.

\subsection{Test rig design}
\label{SMtext:testRig}
Figure \ref{SMfig:tensileTestSetup} is the setup for tensile tests, which is used as an example to show the typical structures of test rigs. We customized a rigid support frame constructed from aluminum T-slotted rails (80 mm $\times$ 80 mm, 80/20) and connectors. This frame was securely attached to the base of the Instron using bolts, ensuring the rigidity of the test rig. Due to the limited space within Instron, we had to redirect the tension through a low-friction pulley while the beam was orientated vertically to align with the moving direction of the Instron. Thus, the force stand was solidly fixed onto a horizontal rail of the frame with its push head connected to the nylon string of 380 mm. Then the test sample was assembled onto the test rig with the top bead fixed on a customized 3D printed connector and end bead attached onto the base. During the pretension state, the push head was moved to certain position to reach desired tension values along its rail and then fixed during the testing to maintain the tension. The test rigs for bending and compressive tests share most of the structures with those for the tensile tests. More details can be found in Fig. \ref{SMfig:bendingTestrig} and \ref{SMfig:TandCTest}.

\subsection{Calibration of Young's modulus}
\label{SMtext:CalibratModulus}

\bigskip
\noindent\textbf{Nylon string}\\
A 380 mm-long nylon string with a diameter of 0.55 mm was attached to the gripper of the Instron machine through two customized connectors (Fig. \ref{SMfig:nylonTest}). The connectors were used to minimize the string's stress concentration caused by sharp edges of Intron's gripper to avoid early breakage. The tests were run using displacement control at a loading rate of 1 mm min\textsuperscript{-1}.

\bigskip
\noindent\textbf{Resin}\\
To obtain the Young's modulus of the resin, we printed a solid cylindrical resin sample with both diameter and height of 15 mm. During test, the sample was placed on the Instron (Fig. \ref{SMfig:resinTest}) with a loading rate of 100 N min\textsuperscript{-1} to obtain the force-displacement curve. This curve was documented after several conditioning cycles to ensure data stabilization.

\subsection{Determination of a string material model for use in FE model}
\label{SMtext:Ogden}
To create a suitable material model for the string, we collected uniaxial force-displacement data on a universal testing machine (Fig. \ref{SMfig:nylonTest}). Using ABAQUS CAE software, we fit the data via a nonlinear least-squares procedure to several hyperelastic material models ({Fig. \ref{SMfig:abaqusOgdenCurveFit}). The Ogden material model with a $N$ value of 3 was most stable and accurate in the displacement range of interest. The strain energy potential in the Ogden form is

\begin{equation}
    U = \sum_{i=1}^{N} \frac{2 \mu_i}{\alpha^2_i} \left(\overline{\lambda}_1^{\alpha_i} + \overline{\lambda}_2^{\alpha_i} + \overline{\lambda}_3^{\alpha_i} - 3 \right) + \sum_{i=1}^{N} \frac{1}{D_i} \left(J^{el} - 1 \right)^{2i}
\label{eq:eq_Ogden}
\end{equation}

\noindent where $N$, $\mu_i$, $\alpha_i$, and $D_i$ are material parameters. $J^{el}$ is the elastic volume ratio and $\overline{\lambda}_i$ are the deviatoric principal stretches. The initial shear modulus is
   
\begin{equation}
    \mu_0 = \sum^{N}_{i=1}\mu_i 
\label{eq:shearModulus}
\end{equation}

The determined values for the material parameters are given in Table \ref{table:ogden}. All values of $D_i$ are zero, meaning that the string material is modeled as incompressible.

\subsection{Weight sustaining of the $2 \times 2 \times 2$ cubic lattice}
\label{SMtext:weightSustaining}
This system requires an initial energy input to deploy the lattice, but once deployed, it locks into its configuration. This means that it does not require a constant energy source. The lattice can be deployed with weights or motor-driven cables to take advantage of this property. We can calculate the energy input into the system in the process of deployment. Because it is path-independent, we can simply calculate the energy of the system after deployment. There are two types of energy to be considered in the deployed lattice: strain energy and potential energy. Strain energy is the energy stored in the compression of the beam due to driving tension. This takes the form
\begin{equation}
    U=V\frac{\sigma^2}{2E}
\end{equation}
where $\sigma$ is the strain, E is the Young's modulus, V is the volume of the beads, respectively. For a beam with length L and cross-sectional area A, this reduces to
\begin{equation}
    U=\frac{LF^2}{2EA}
\end{equation}
where F is the tensile force applied to the string. 
For our lattice, the strain energy is estimated as 0.32 J. The other energy is the potential energy, which is the energy to lift the beads off of the ground into the deployed state. This has the formula $P=mgh$, where h is the height of the center of mass. We calculate that by averaging the heights of all the beams in the lattice, which gives us potential energy about 0.15 J for a total energy input of $\sim$0.47 J.

\subsection{Discussion on potential applications}
\label{SMtext:applications}
Our system’s self-deployability makes it well-suited for designing machinery used in space while the tunability of mechanical properties is particularly advantageous for dynamic environments \cite{pellegrino2001deployable}. For example, the flexibility of our undeployed material could simplify packing large lightweight structural lattices into tight and irregular spaces like a cargo bay on a small rocket. Once deployed, our material would allow these lattices to alternate between stiffening to support high loads during maneuvers and relaxing to dissipate latent vibrations from the rocket thrust.  Our system can also be used to build robots for navigating complicated terrains \cite{baines2022multi}. For example, stiffness can be increased for fast movement over even surfaces, while it can be softened for traversing rough terrain or for sustaining impact (e.g., air-dropping, landing, and collision)\cite{yan2015experimental,hong2015experimental}. Similarly, our system can also be used in human-machine interaction \cite{mintchev2019portable} and medical devices\cite{wang2022adaptive}, where self-deployability and tunability of mechanical properties are necessary.

\subsection{Theoretical model}
\label{SMtext:theoretical model}

This code describes a geometry-based model for calculating the forces and displacement on a series of beads held by a string in tension. When the beads are in contact, the system is fully defined by the local rotation and relative displacement, allowing us to calculate other interactions using the laws of sine and cosine. As Fig. \ref{fig:schematicBead} shows, there are three distinct modes of contact between pairs of beads: surface contact, where the faces are flush with each other; two-point contact, which occurs when the bead angle $\theta$ is small; and one point contact, which occurs with a large bead angle or rotation angle. Each condition provides a complete system of equations to solve for the static deflection, as the three force-balance equations ensure contact and solve for displacement and rotation. Starting at the free end, with the attached string force on the right end, you can calculate the applied contact force from the preceding bead until you get to the base. An important assumption is that all the forces are applied perfectly normal to the surface, without deformation. For each contact source, there are two input forces from the previous bead, and a force and moment to balance.

In each iteration of the loop, the deflection force is increased slightly from the previous configuration, and the initial forces in the x and y direction on the bead are calculated. Then, based on these values, the reaction forces are calculated. The resulting moment is used to move the bead, and the process is iterated until it converges. However, the forces are not continuous as the contact type changes, especially as it moves from surface to two-point contact. This is because the locus of the force changes from the center of the surface to the edge in contact in the two-point region. As a result, the model moves stepwise from stable configuration to stable configuration.

Each bead is acted on not only by the adjacent beads, but also by the string, as it is bent around the edge of the internal hole when the beads are unaligned. Static friction is also added, which has the effect of increasing the range of stable configurations before sliding is induced. By starting at the free end, the forces on each bead can be solved sequentially, never requiring more than two unknown variables.

\noindent \textbf{\large Parameters}\\
The design of the bead took in 8 parameters, shown in Table \ref{tab:parametersOfBeads}.

\noindent \textbf{\large Bead model}\\
The bead is a rigid body defined by two parallelograms, mirrored across an axis. The code uses the 8 points on the corners, which is enough to fully define the object. The relevant parameters are $R$, the outer radius, $r$, the inner radius, $\theta$, the bead angle, $w$, the length of the bead, and $\mu$, the friction coefficient (Fig. \ref{fig:schematicBead} and Fig. \ref{fig:beadModel}). 

The string is defined by the spring coefficient, which can be either constant or a function of length. 
Although the system being modeled is 3-dimensional, the cross-section at the midline almost fully describes the behavior. During one and two point contact, the contact point is in this plane, and by symmetry no force acts out of the plane. During surface contact, there is no movement until the force is concentrated at the bottom tip of each face, which also lies on the centerline. Therefore, we do not need to consider the system in the out-of-plane dimension.

In the code, the position of the bead is the location of the front tip, where the two front faces would intersect.  This can be calculated from its (lower) contact point $\vec x_p$, rotation $\phi$ and displacement from the bottom corner $y$ relative to the prior bead. The horizontal displacement is calculated to maintain contact. For a positive displacement, which is defined in the $-y$ direction, the bottom corner $\vec x_c$ is located at 
\begin{equation}
    \vec x_c=\vec x_p-y(cos(\theta - \phi),sin(\theta - \phi))
\end{equation}
For a negative displacement, the bottom corner is the contact point. This occurs for very small bead angles, and some one point contact scenarios. From this corner at $\vec x_c$, the bead tip $\vec x_t$ is calculated as 
\begin{equation}
    \vec x_t=\vec x_c+w(cos\phi,sin\phi)+R(cos\phi,sin\phi tan\theta)
\end{equation}

\noindent \textbf{\large Assembly}\\
The bead is a class, with the parameters that define the shape along with the position of the the angle between it and the previous bead $\phi$ and the displacement between the bottom corners $d$. A useful intermediate parameter $pos$, the location of the front tip of the bead, is calculated from the angle and displacement as well. This procedure is done iteratively, starting from the base. This can assemble the beads with any intial angle and displacement. The initial bead is set with pos at $(0,0)$, with no angle or displacement. The next bead has point $\vec G$ calculated according to
    \begin{equation}
        \vec G=\vec x_c-d(cos(\theta-\phi),sin(\theta-\phi))
    \end{equation}
    if the displacement is positive (point $\vec H$ on the previous bead is above point $\vec G$ on this bead) and 
    \begin{equation}
        \vec G=\vec x_c-d(cos(\theta),sin(\theta))
    \end{equation}
    if the displacement is negative. $\vec H$ is then calculated for this bead by adding $w(cos\phi,sin\phi)$ to point G, and $\vec pos$ is calculated by adding $\frac{R}{sin\theta}(cos(\theta-\phi),sin(\theta-\phi))$. With $\vec pos$ and $\phi$, all the other points that define the bead can be calculated easily. The beads are initialized like this until you have $n_{beads}$.

\noindent \textbf{\large String model}\\
The string is initialized next. It is two arrays, each with $2*(n_{beads}+2)$ elements for each opening of each hole in the system. One array is the position of the point, and the other is the location of the contact (top, bottom, or none). To determine whether there was contact, first it was determined whether the string was passing through or out of a bead. Then, the intersect function was run, which calculated the current and next possible positions (at either side, or in between the internal radius). The three relevant values were the initial value $x_0$, the proximate entry or exit $x_p$, and the subsequent exit or entry $x_s$. If both points on the bead at $x_p$ were above the line between $x_0$ and $x_s$, there was contact at the bottom of the bead. Similarly, if both outs on the bead at $x_p$ were below the line between $x_0$ and $x_s$, there was contact at the top of the bead. If neither of these were true, there was no contact at that location. In that case, the next intersection point was determined, and all intermediate values were set to 'none'. At the end, the distances are summed to calculate the tension.
The string running through the beads is under tension, so when it is bent around a corner, it applies a force against the bead. This force can be calculated by creating a force balance at the point of contact. The angle in the string can be calculated from the adjacent string points. Assuming constant tension $T$ and an angle of $2\phi$ in the string, the force is $2*Tcos\phi$ applied along the line of symmetry. Alternatively, you can do a vector sum of the tension forces in the string from that point to get the same answer. Whenever the system moves, a check is run to determine where the string is in contact, by checking if the string can pass cleanly from the point behind it to the point in front. The tension in the string is nonlinear and interpolated from a table of experimental values.

\noindent \textbf{\large Contact forces}\\
\noindent \textbf{Surface contact.}
When the beads are aligned with each other, the forces are assumed to be normal to the surface. The forces from the string and the bead to the right are applied, and the force sum in the x and y directions are calculated. Because the angles of the forces are fixed, only the magnitudes need to be solved for, giving two equations and two variables. For $F_x$ and $F_y$, our system is
\begin{equation}
    F_x=F_{top}cos\theta+F_{bot}cos\theta
\end{equation}
\begin{equation}
    F_y=F_{top}sin\theta-F_{bot}cos\theta
\end{equation}
$F_{top}$ and $F_{bot}$ are calculated implicitly. Then, from this the moment is calculated as
\begin{equation}
    M=\Sigma FxsinF_{\theta} - FycosF_{\theta}
\end{equation}

for forces $\vec{\textbf{F}}(F,F_{\theta},\vec{\textbf{x}})$, where $F$ is the magnitude, $F_{\theta}$ is the angle, and $\vec{\textbf{x}}$ is the position. The location of the forces are moved to balance this out, according to 

\begin{equation}
\begin{aligned}
    M=&F_{top}(y-qsin(\theta+\phi))cosF_{\theta} - F_{top}(x+qcos(\theta+\phi))sinF_{\theta} \\
    &+F_{bot}(y-qsin(\theta+\phi))cosF_{\theta} -F_{bot}(x-qcos(\theta+\phi))sinF_{\theta}
\end{aligned}
\end{equation}
with q as the distance along the edge from the tip. Once it reaches the bottom and the moment is still negative, the bead is rotated proportionally based on this value, and the process iterates, moving to one or two point contact based on the bead angle. When the angle is smaller than 45 degrees, it is initially in two point contact, while it is in one-contact if the angle is larger.

\noindent \textbf{Two-point contact.}
When the beads are in two point contact, the forces are applied at the points of contact and normal to the surface again. These have angles normal to the bead with rotation $\phi$, and are located at the contact point $\vec x_p$ and the front inside corner of the top bead. The force sum components from these vectors are

\begin{equation}
    F_x=F_{top}cos(\theta-\phi)+F_{bot}cos(\theta)
\end{equation}
\begin{equation}
    F_y=F_{top}sin(\theta-\phi)-F_{bot}sin(\theta)
\end{equation}
The same force balances are calculated, and the bead is rotated based on the moment. This movement is fully defined while the system is in two point contact, as the two faces slide over each other. For a given rotation, the maximum displacement (measured as the distance along the lower side between the bead corners) is 
\begin{equation}
    disp=L-sin(\pi-\phi-2*\theta)*L/sin(2\theta)
    \label{eq:dispBead}
\end{equation}
where L is the length of the bead face, $\phi$ is the angle between the beads, and $\theta$ is the bead angle. This is calculated from the law of sines on the triangle of the two contact points and the crook of the bead in front.

\noindent \textbf{One-point contact.}
When the beads are in one point contact, the force location is fixed to the same point. The two forces are the normal and friction force, with angles
\begin{equation}
    F_x=F_{norm}cos\theta+F_{fric}cos(\theta-\frac{\pi}{2})  
\end{equation}

\begin{equation}
    F_y=F_{norm}sin\theta-F_{fric}sin(\theta-\frac{\pi}{2})
\end{equation}
The friction force is limited to a given fraction of the normal force, based on the friction coefficient. Therefore, it is possible to not be able to sum to 0 in the y direction. This causes the bead to slide downward, while the moment still causes rotation. This gives one point contact two degrees of freedom for motion. Through this movement, it can return to two point contact if the downward displacement is large. The resulting minimum angle is calculated using the law of cosines for the triangle defined by the two sides and the inside angle of the bead. This gives
\begin{equation}
    \phi=2 \theta -arcsin(L-disp)*sin(2\theta)/L)
    \label{eq:phi}
\end{equation}

\noindent \textbf{Solver.}
Once all the forces and moments were calculated, the beads were adjusted moving from the free end to the fixed end. Each bead was rotated an amount that scaled linearly with the calculated moment, and this was applied to all beads after it as well to preserve their relative angle. If, after this, a bead was found to have a rotation less than the prior bead, it was set to the prior bead's angle and the contact type set to surface. All of the changes to displacement and angle were larger for the first hundred timestep, before being reduced to 1/10 of their value for precision.

For one point contact, both the $disp$ and $\phi$ were changed based on the force in the y direction and moment, respectively. Then, it was checked if the beads were in two point contact by checking the minimum angle and displacement functions as described above. If it was two point contact, the displacement was fixed according to Eq. \ref{eq:dispBead}. If the force at the top point of contact was zero, the contact type was switched to `one'. Finally, if the force sum in the y direction was not zero, the contact type was set to `two', the displacement was calculated from the force, and the two point contact angle from Eq. \ref{eq:phi} was applied. Under surface contact, the contact type is changed when the forces are applied beyond the edges.
A sample beam with one point contact for the first bead, two point contact for the next five and surface contact for the remaining (Fig. \ref{fig:theoreticalResult}).

\subsection{Comparing experimental, numerical, and theoretical results.}
\label{SMtext:compare}
A theoretical model can provide valuable insights for the deflection of beams constructed with high stiffness beads (Fig. \ref{fig:theoreticalResult}). To more generally explain the new metamaterial behaviors discovered in our CCPJ system, we decided that we could provide the clearest characterization using finite element techniques validated by empirical data, rather than presenting a more complex, more opaque analytical model.

In Fig. \ref{fig:comparison}, we show the experimental bending force-displacement curve at an initial tension of 50N for a beam with 40-degree beads alongside the disjoint curve created by the corresponding analytical model. We also provide the force-deflection curves of finite element models having a range of elastic moduli assigned to the beads. 

Our simplified theoretical model assumes perfectly rigid beads and thus could be used as an upper bound when optimizing structures with high elastic modulus bead materials. While the experimental results are characterized by a smooth transition between high and low stiffness behavior, the analytical model shows an abrupt breaking point in the force-deflection behavior. At this point, the beam undergoes a dramatic decrease in stiffness. Through finite element studies, we see that this breaking point represents an approximate limit to the deflection force a beam can sustain at high stiffness. When beads are made of a low elastic modulus material, the FE force-deflection behavior reflects the experimental results of a beam with low elastic modulus beads. As the bead elastic modulus increases, the FE force-deflection behavior approaches that of the analytical model; it develops a high initial stiffness and then transitions to a low stiffness at a breaking force. The analytical model, having perfectly rigid beads, approximates the upper limit of that breaking force.

The finite element models allow elasticity to be considered in conical bead geometries. Implementing this capability in a theoretical model would be complex as it involves nonlinear deformation in multiple dimensions. The FE studies therefore play a key role, allowing us to vary the bead’s elastic modulus and discover that the modulus has a major effect on the force-displacement behavior of the CCPJ beams.

\subsection{Effect of length of beads on the mechanical properties.}
\label{SMtext:effectLength}
The length of the bead is an important parameter that affects the properties of the proposed metamaterials. We have run preliminary finite element simulations to show how the initial beam stiffness is affected by varying the individual bead length (Fig. \ref{fig:FEALength}). The total length of the unit beam is kept constant at 150 mm of unfixed length and the deflection contact is kept at 7.5 mm from the front edge of the cylindrical outer surface of the first bead. Five different scenarios are run: 15 bead (10 mm length), 10 bead (original 15 mm length), 8 bead (18.75 mm length), 6 bead (25 mm length), and 3 bead (50 mm length). These preliminary results indicate that the stiffness of the beam increases as the length of the beads increases (Fig. \ref{fig:effectLength}). In addition, as the reviewer points out, the flexibility of the proposed system decreases as the beads get longer. Thus, for practical applications, a designer must consider the desired stiffness vs flexibility when deciding on a bead length.

\clearpage

\subsection*{Supplementary Figures}

\begin{figure*}[ht]
  \centering
  \includegraphics[trim=1in 12.3cm 1in 0.8cm, clip=true,width=1\textwidth]{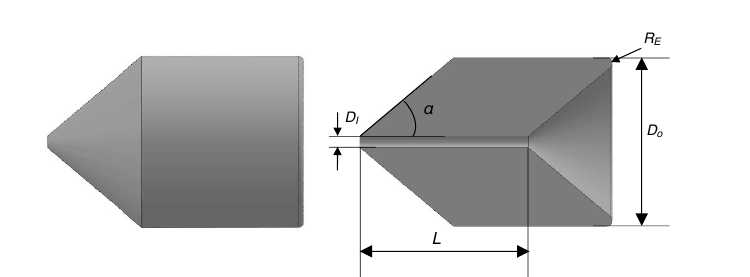}
  \caption{\textbf{Geometry of a proposed bead with matching conical concavo–convex interfaces.} $D_O$ is the outer diameter, $D_I$ is the inner diameter, $L$ is the length, and $\alpha$ is the angle of the cone. $R_E$ is the radius of the edge. Unless specified otherwise, $R_E$ is set to a default value of 0 mm for fabrication.}
  \label{SMfig:beamGeometry}
\end{figure*}

\begin{figure*}[ht]
  \centering
  \includegraphics[trim=0in 0cm 0in 0cm, clip=true,width=0.45\textwidth]{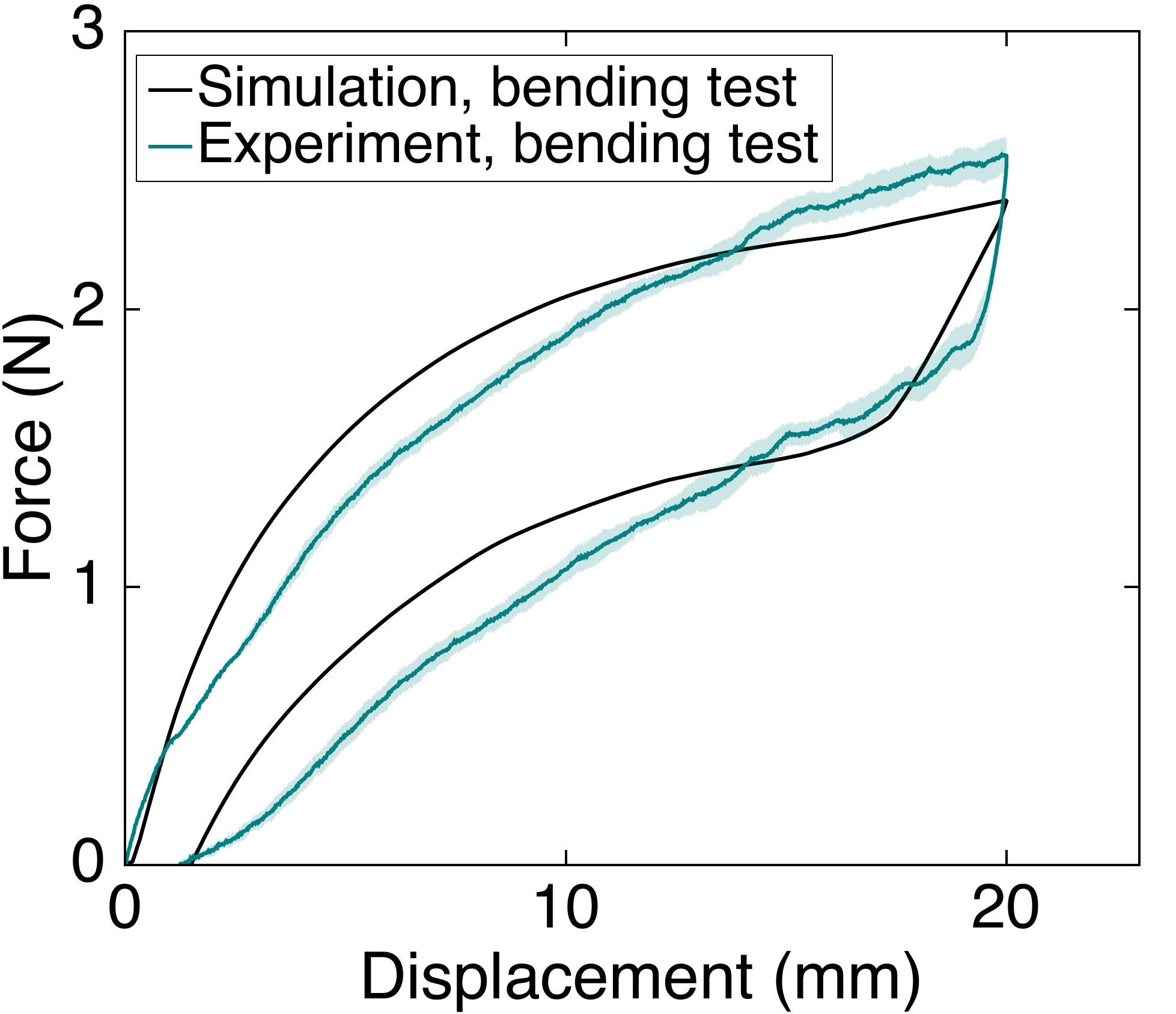}
  \caption{\textbf{Comparison of the force-displacement curves between the FE model and experiment tests for a beam with 40$^\circ$ beads.} The string is pretensioned at 50 N. The shaded area represents the standard deviations between three different experimental tests.}
  \label{SMfig:ComparisonModelFEA}
\end{figure*}

\begin{figure*}[h!]
  \centering
  \includegraphics[trim=0in 0cm 0in 0cm, clip=true,width=0.8\textwidth]{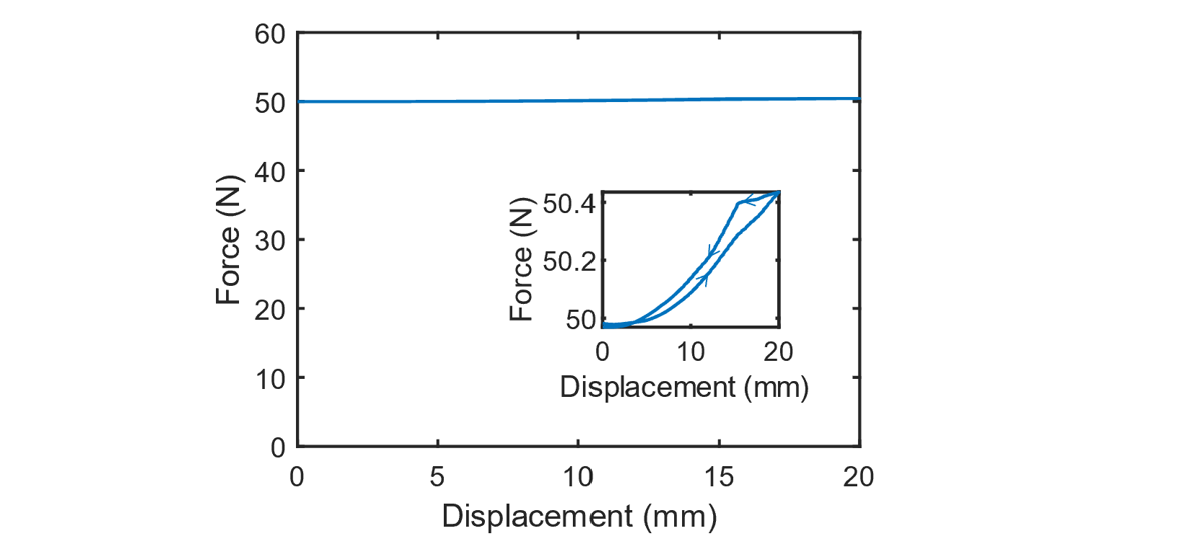}
  \caption{\textbf{Tension variance of the string during a FE bending indentation.} The tension is recorded during a FE bending study of a beam with 40$^\circ$ beads after the string is initially tensioned to 50 N and then fixed in place at the free end. The displacement on the horizontal axis refers to the distance traveled by the indenter.}
  \label{SMfig:tensionVarianceOfString}
\end{figure*}

\begin{figure*}[ht]
  \centering
  \includegraphics[trim=0in 6.4cm 0in 0.62cm, clip=true,width=1\textwidth]{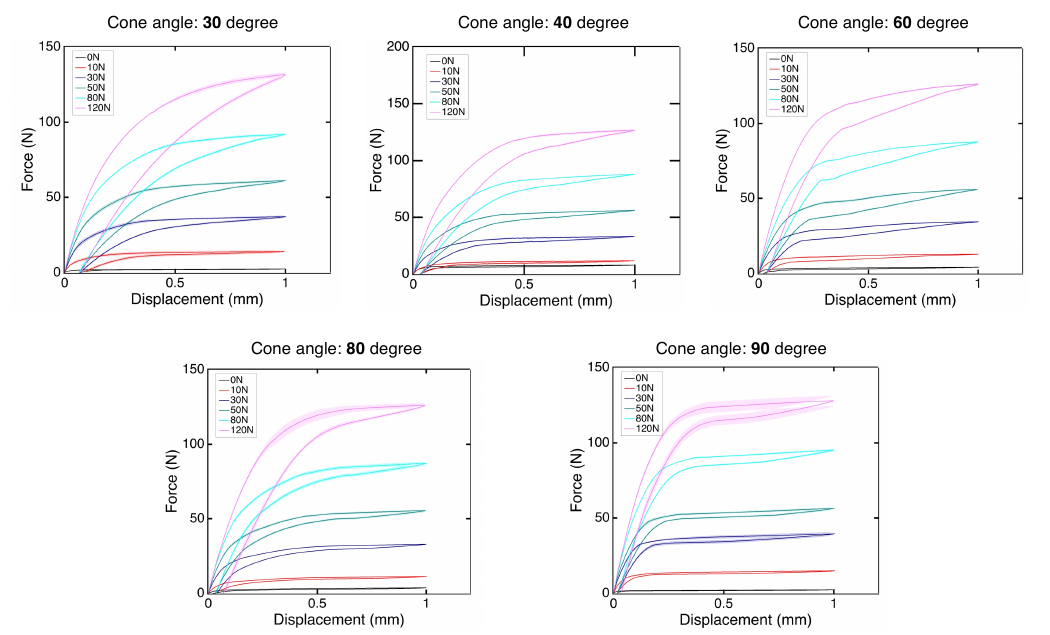}
  \caption{\textbf{Force-displacement curves of tensile testing as a function of contracting tension over different cone angles.} The shaded areas represent the standard deviations between three different experimental tests.}
  \label{SMfig:tensileTestOverAngles}
\end{figure*}

\begin{figure*}[ht]
  \centering
  \includegraphics[trim=0in 6.4cm 0in 0.62cm, clip=true,width=1\textwidth]{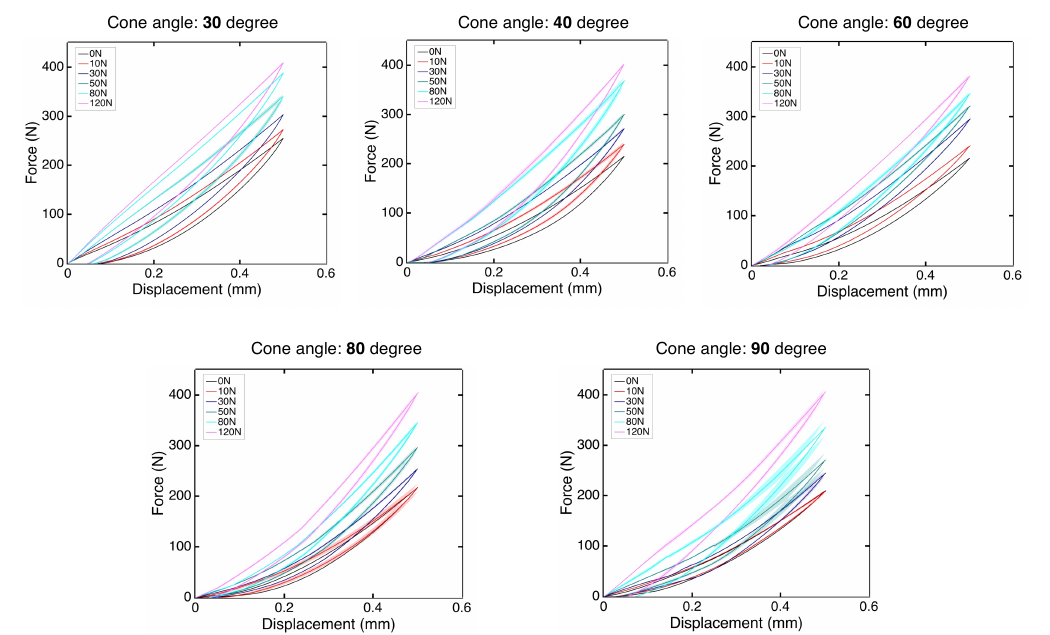}
  \caption{\textbf{Force-displacement curves of compressive testing as a function of contracting tension over different cone angles.} The shaded areas represent the standard deviations between three different experimental tests.}
  \label{SMfig:compressiveTestOverAngles}
\end{figure*}

\begin{figure*}[!htb]
  \centering
  \includegraphics[trim=1in 12.4cm 1in 0.6cm, clip=true,width=1\textwidth]{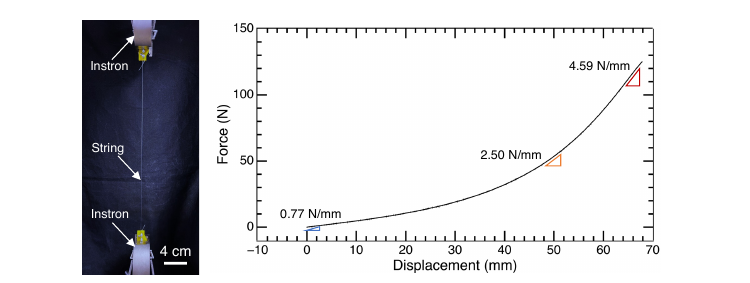}
  \caption{\textbf{Nylon tensile test.} A 380 mm-long nylon string with a diameter of 0.55 mm was tested with its two ends attached onto Instron machine. The force-displacement curve was obtained after several training cycles.}
  \label{SMfig:nylonTest}
\end{figure*}

\begin{figure*}[ht]
  \centering
  \includegraphics[trim=1in 12.2cm 1in 0.6cm, clip=true,width=1\textwidth]{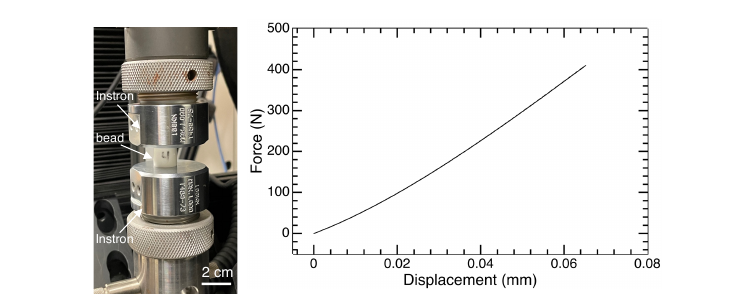}
  \caption{\textbf{Resin compressive test.} A cylindrical resin sample with both diameter and height of 15 mm was placed on the Instron to obtain the force-displacement curve. This curve was recorded after several training cycles to stabilize the data.}
  \label{SMfig:resinTest}
\end{figure*}

\begin{figure*}[ht]
  \centering
  \includegraphics[trim=0in 6cm 0in 0.62cm, clip=true,width=1\textwidth]{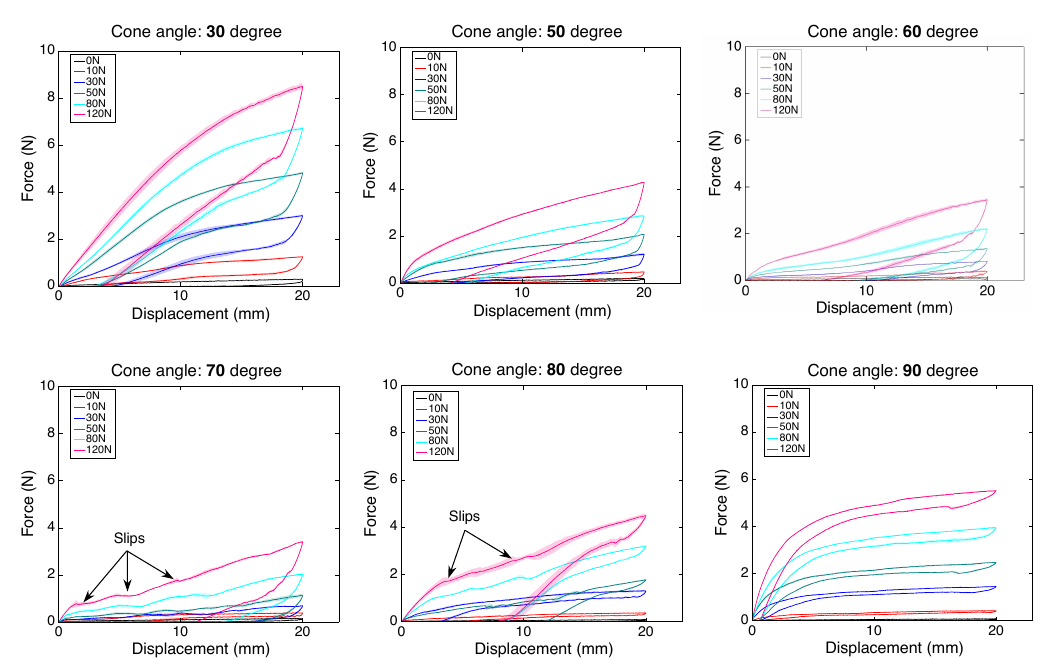}
  \caption{\textbf{Force-displacement curves of bending testing as a function of contracting tension over different cone angles.} The shaded areas represent the standard deviations between three different experimental tests.}
  \label{SMfig:bendingTestOverAngles}
\end{figure*}

\begin{figure*}[ht]
  \centering
  \includegraphics[trim=0.4in 93mm 0.4in 7mm, clip=true,width=1\textwidth]{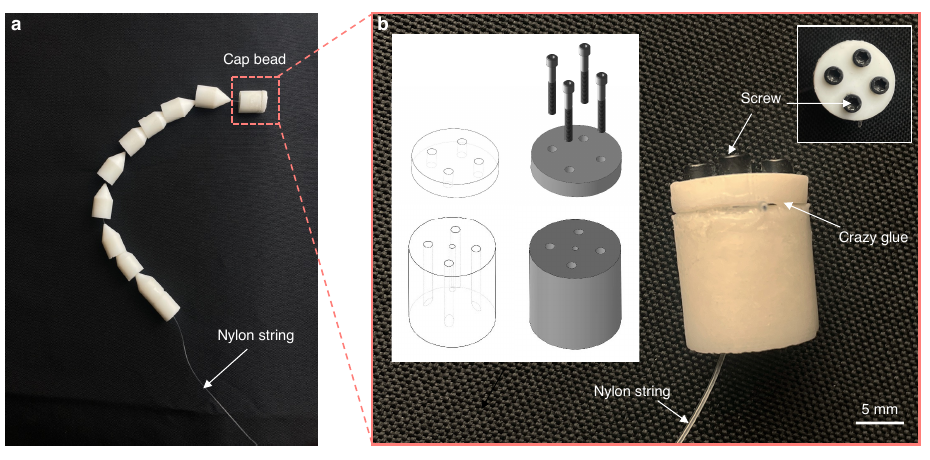}
  \caption{\textbf{Assembly of a beam.} \textbf{a}, Image of an assembled beam with 11 beads. \textbf{b}, The assembly of the top bead. A nylon string was firstly fed through the central hole of the top bead and clamped onto it with four screws. To secure the connection between the string and top bead, crazy glue was also used, which can sustain more than 120 N pulling force on the string.}
  \label{SMfig:beamAssembly}
\end{figure*}

\begin{figure*}[ht]
  \centering
  \includegraphics[trim=1in 12.5cm 1in 5mm, clip=true,width=1\textwidth]{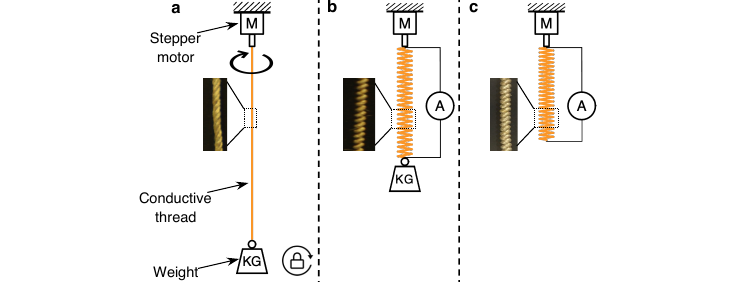}
  \caption{\textbf{Fabrication of the super-coiled polymer actuators (SCPAs).} \textbf{a}, Implementation of coils. Under a pretension created by a hanging weight, the conductive yarn was twisted using a stepper motor to form coils. \textbf{b}, Coiled yarn annealing. The stress in the coiled yarn was relieved through a cyclic heating and cooling process, transforming it into a CSCP actuator. \textbf{c}, Actuator Stabilization. Post weight removal, the actuator underwent a similar thermal treatment to stabilize the strain.}
  \label{SMfig:SCPA}
\end{figure*}

\begin{figure*}[ht]
  \centering
  \includegraphics[trim=1in 11.5cm 1in 5mm, clip=true,width=1\textwidth]{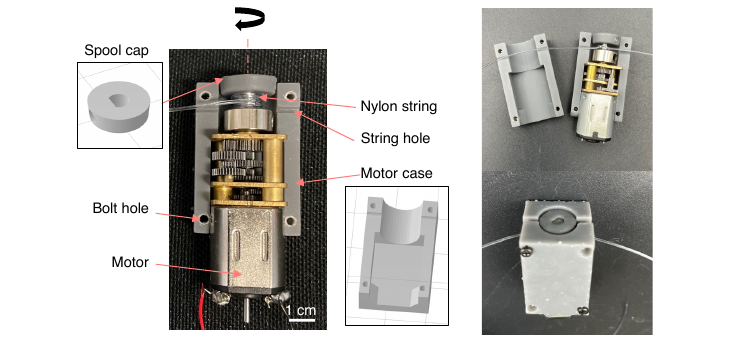}
  \caption{\textbf{The design and assembly of a motor-driven cable (MDC) module.} A customized case was used to host the motor. A nylon string was pinched between the spool and shaft of the motor through a set screw to improve the connection strength. When powered, the shaft starts to rotate to shorten the nylon string. Owing to the small spool diameter, the MDC can output maximum tension up to $\sim$140 N. This output tension is sufficient for most applications in this paper. }
  \label{SMfig:MDC}
\end{figure*}

\begin{figure*}[ht]
  \centering
  \includegraphics[trim=1in 9.8cm 1in 1cm, clip=true,width=1\textwidth]{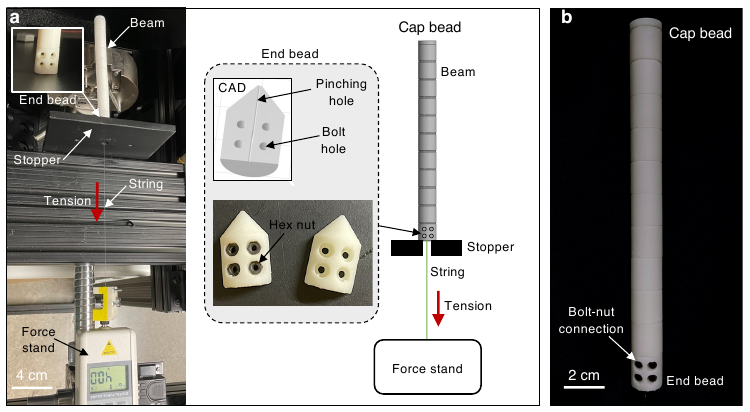}
  \caption{\textbf{Assembly of a beam for a unit cell lattice.} \textbf{a}, The assembly setup and process. \textbf{b}, An example beam assembled with pretension of 40 N on the string.}
  \label{SMfig:beamFabLattice}
\end{figure*}

\begin{figure*}[ht]
  \centering
  \includegraphics[trim=0.5in 7.8cm 0.5in 1cm, clip=true,width=1\textwidth]{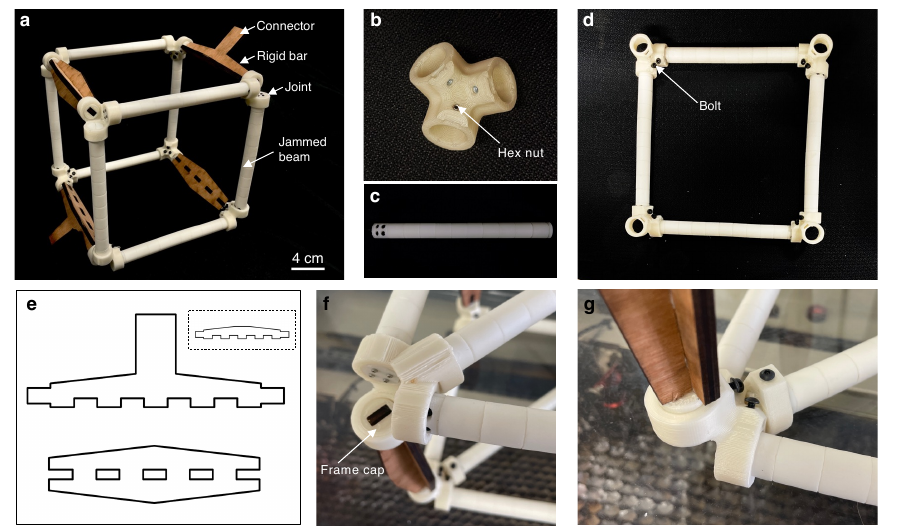}
  \caption{\textbf{Cubic unit cell assembly.} \textbf{a}, The image of a bending-dominated lattice. \textbf{b}, The image of a joint with three Nex nuts, which are used along with bolts to fix beams. \textbf{c}, An assembled beam. \textbf{d}, Four beams are assembled into a square shape with four joints. Terminal beads of each beam are pinched on the joints by bolts. \textbf{e}, The 2D CAD files depict rigid bars designed for laser cutting. \textbf{f} and \textbf{g}, Zoom-in images of the assembled lattice, showing detailed connections.}
  \label{SMfig:unitAssembly}
\end{figure*}

\begin{figure*}[ht]
  \centering
  \includegraphics[trim=0.5in 9.3cm 0.5in 0.8cm, clip=true,width=1\textwidth]{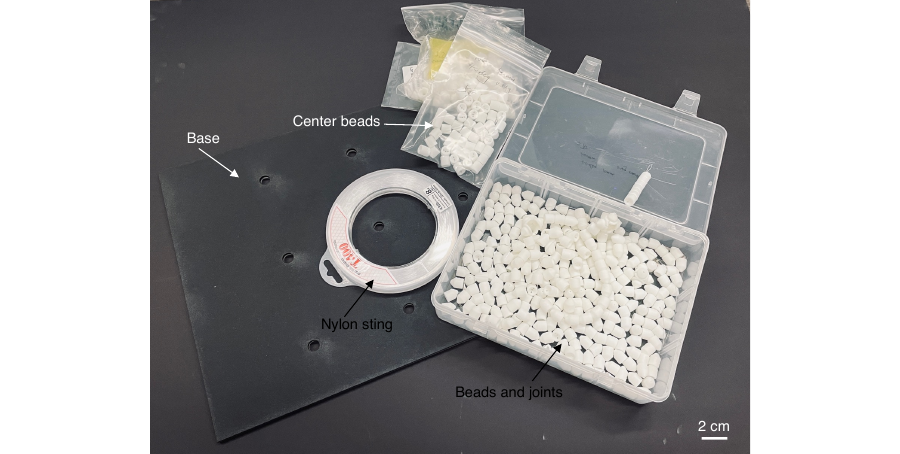}
  \caption{\textbf{Materials for the $2 \times 2 \times 2$ cubic lattice.} The materials include 3D printed hollow beads (and center beads), a nylon string, a rigid base, and customized resin joints.}
  \label{SMfig:materialsLattice}
\end{figure*}

\begin{figure*}[ht]
  \centering
  \includegraphics[trim=0.5in 7.3cm 0.53in 0.8cm, clip=true,width=1\textwidth]{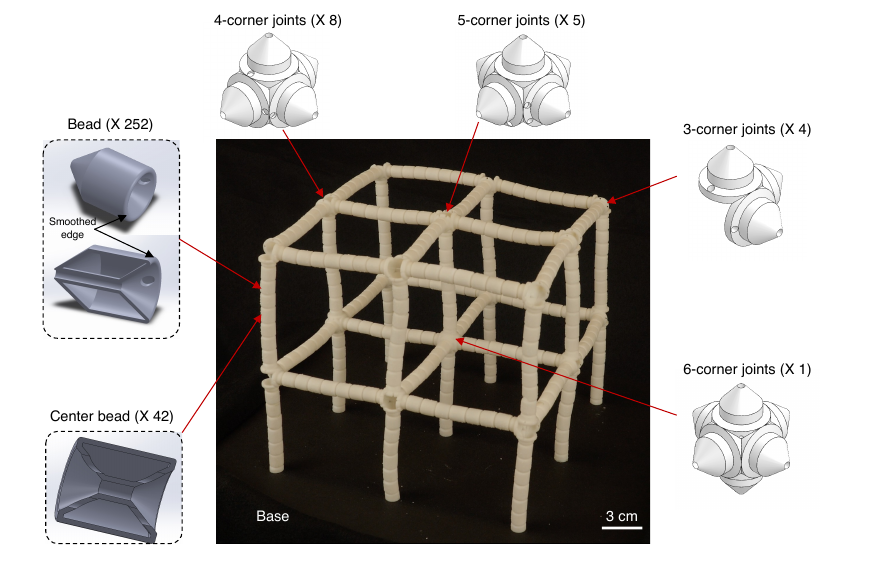}
  \caption{\textbf{Assembly of the $2 \times 2 \times 2$ cubic lattice.} The beams are connected with customized joints. There are in total four different joints, i.e., 3-corner joints, 4-corner joints, 5-corner joints, and 6-corner joints. For example, each 3-corner joint has three corners that are designed as cone shapes to interface with beads. These joints are assigned to different locations, as shown in the image. The base, a laser-cut acrylic board, is used to support the whole assembly. The required number of each component is noted in the parentheses. Several nylon strings were routed through all beads for actuation (Fig. \ref{SMfig:stringPath}).}
  \label{SMfig:assemblyLattice}
\end{figure*}

\begin{figure*}[ht]
  \centering
  \includegraphics[trim=0.5in 10.3cm 0.5in 0.8cm, clip=true,width=1\textwidth]{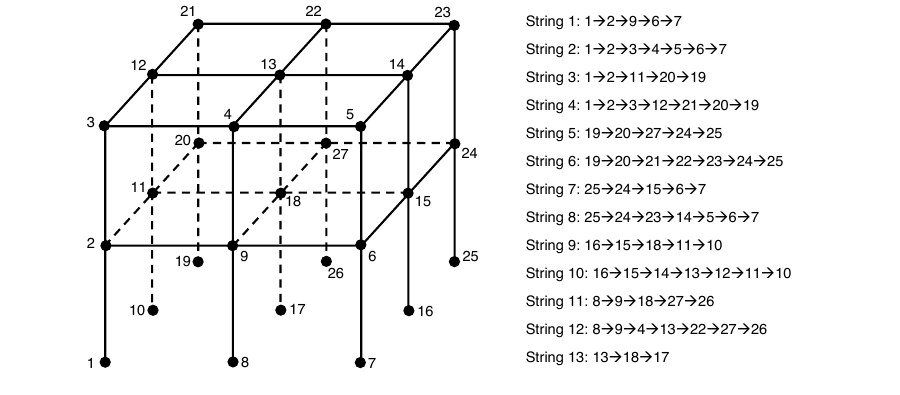}
  \caption{\textbf{String path of the $2 \times 2 \times 2$ lattice.} Each number represents a vertex of the lattice. One end of each string is fixed on the first vertex, while the other is free for actuation. In total, there are 13 string used to go through 27 vertexes. Taking String 1 as an example, one end of the string is fastened onto vertex 1. The other end goes through vertex 2, 9, 6, and 7 subsequently.}
  \label{SMfig:stringPath}
\end{figure*}

\begin{figure*}[ht]
  \centering
  \includegraphics[trim=1in 9cm 1in 0.65cm, clip=true,width=1\textwidth]{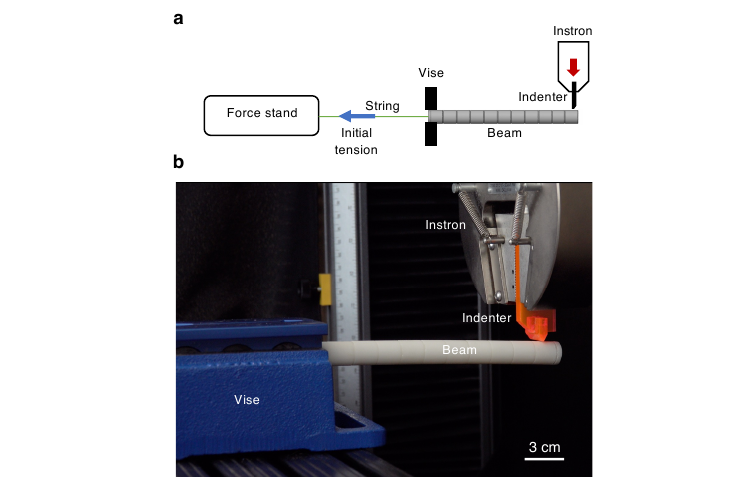}
  \caption{\textbf{Bending testing of a CCPJ-based beam with 40$^\circ$ beads.} \textbf{a}, Testing schematic. \textbf{b}, Image of the test rig. The force stand with its associated supporting frames are not shown here. The indenter is made out of acrylic plate and sufficiently rigid compared to the stiffness of the beam.}
  \label{SMfig:bendingTestrig}
\end{figure*}

\begin{figure*}[ht]
  \centering
  \includegraphics[trim=0.5in 5.6cm 0.8in 0.65cm, clip=true,width=1\textwidth]{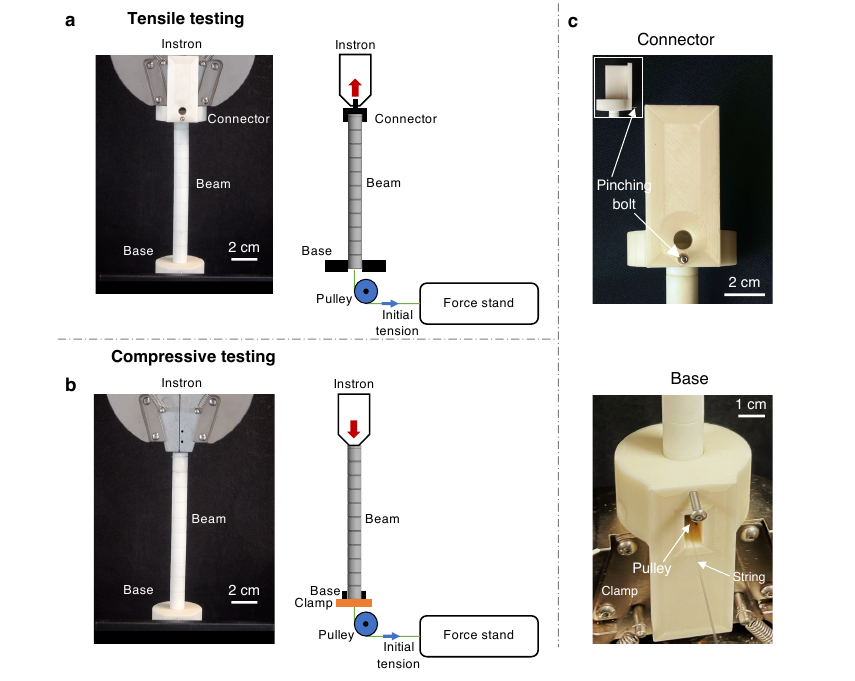}
  \caption{\textbf{Tensile and compressive testing of a CCPJ-based beam with 40$^\circ$ beads.} \textbf{a}, The schematic for tensile testing with labeled experimental image. \textbf{b}, The schematic for compressive testing with labeled experimental image. \textbf{c}, The detailed structure of connector and base.}
  \label{SMfig:TandCTest}
\end{figure*}

\begin{figure*}[!htb]
  \centering
  \includegraphics[trim=1in 13cm 1in 0.8cm, clip=true,width=1\textwidth]{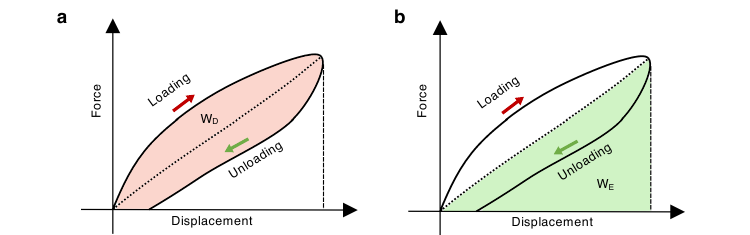}
  \caption{\textbf{The definition of loss factor based on a force-displacement curve.} \textbf{a}, The dissipated energy $W_D$ during the loading-unloading phase. \textbf{b}, Graphical representation of the stored energy $W_E$, which is approximated as the sum of the half of the $W_D$ and the enclosed area of the unloading curve with x-axis \cite{zhang2022dynamic}.}
  \label{SMfig:lossFactor}
\end{figure*}

\begin{figure*}[ht]
  \centering
  \includegraphics[clip=true,width=1\textwidth]{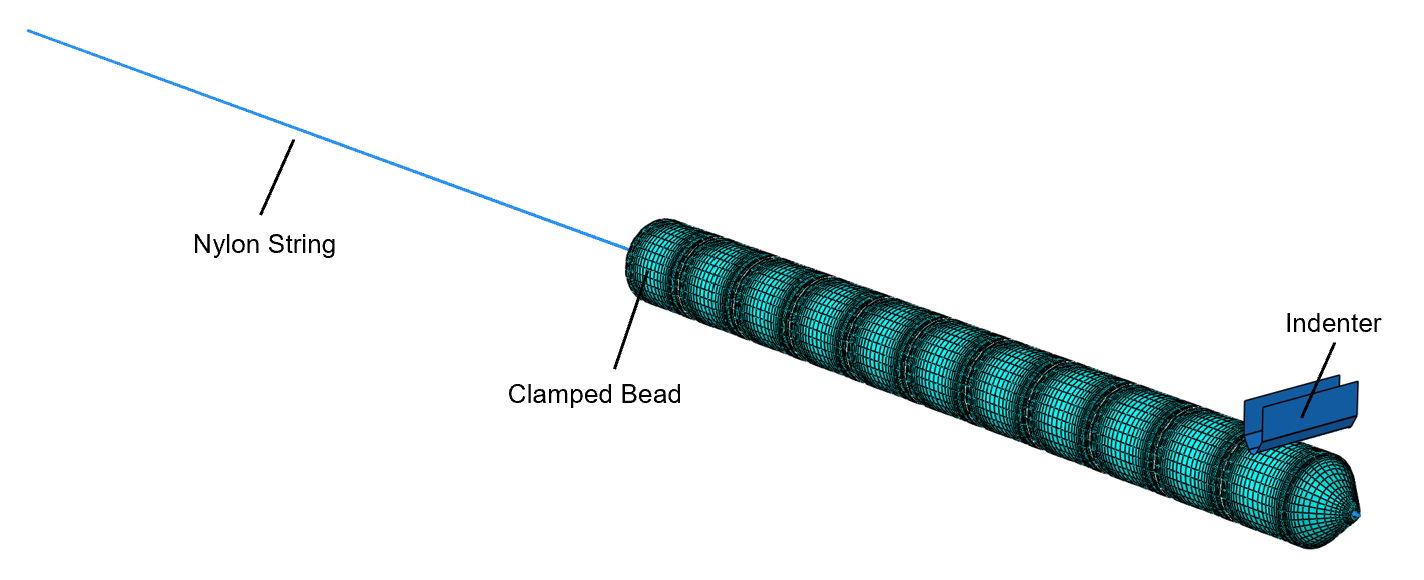}
  \caption{\textbf{Meshed assembly of a CCPJ-based beam with $40^{\circ}$ cone angles as prepared for an FE quasi-static bending analysis in ABAQUS CAE software.} Models were created at $10^{\circ}$ increments for cone angles $30^{\circ}$ through $90^{\circ}$.}
  \label{SMfig:feaWholeAssem}
\end{figure*}

\begin{figure*}[ht]
  \centering
  \includegraphics[clip=true,width=1\textwidth]{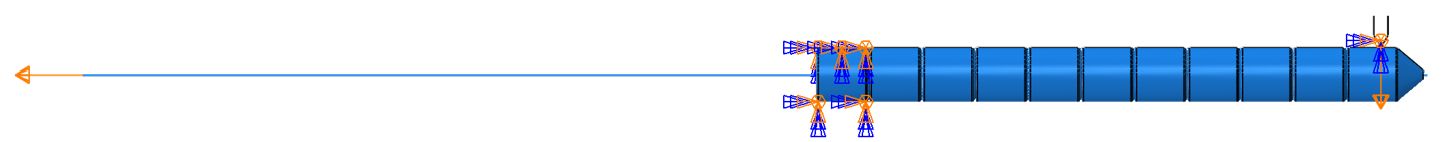}
  \caption{\textbf{Loading conditions for the FE quasi-static bending analysis.} Orange arrows represent prescribed displacements. Blue arrows represent prescribed rotations. All prescribed values are 0 with the exception of the displacement on the left of the string and the vertical displacement on the indenter, which vary as described in Methods.}
  \label{SMfig:feaLoadingConditions}
\end{figure*}

\begin{figure*}[ht]
  \centering
  \includegraphics[trim=2in 2mm 2in 0cm, clip=true,width=0.5\textwidth]{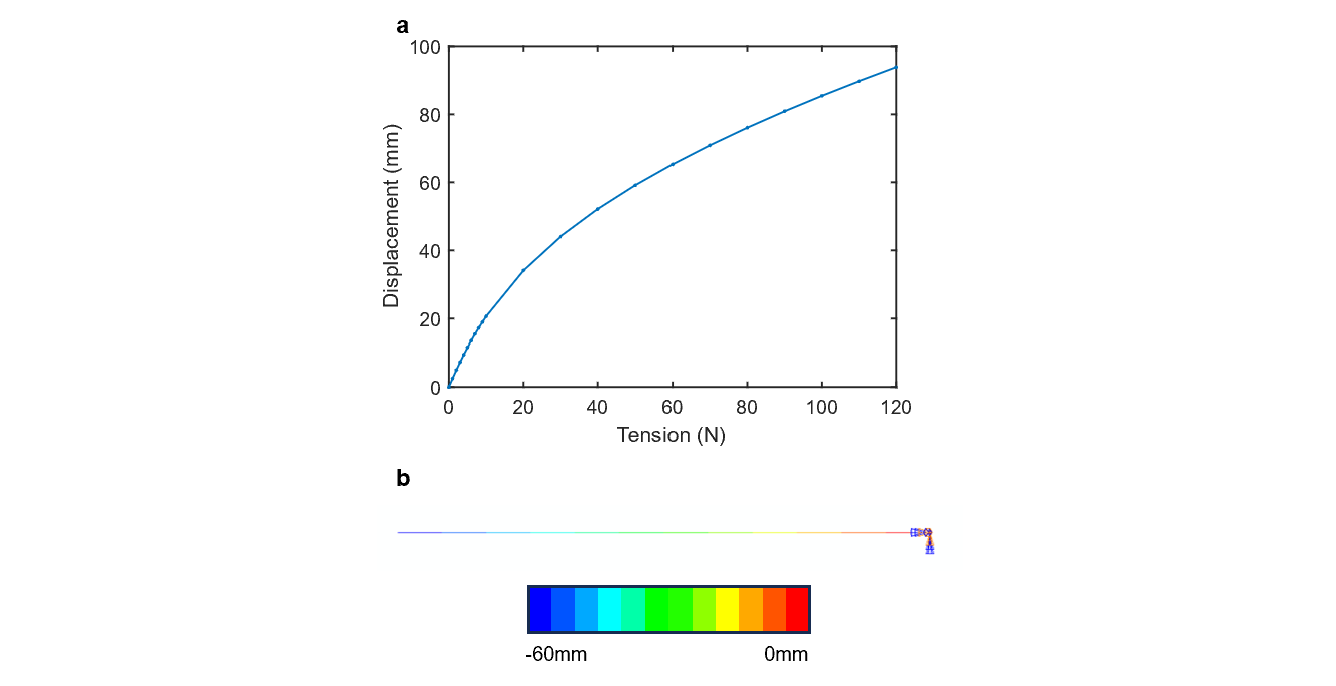}
  \caption{\textbf{Determination of prescribed string end displacement for FE analyses.} \textbf{a}, Plot of the prescribed displacement load to be used as a function of the desired pretension load from 0 N to 120 N. \textbf{b}, Displacement plot of a 380 mm long nylon string with a 50 N tensile force applied to the left end of the string, with the right end being fixed. The same FE analysis was run for all pretension loads of interest.}
  \label{SMfig:feaPretensionPlot}
\end{figure*}

\begin{figure*}[ht]
  \centering
  \includegraphics[trim=0in 11.2cm 0in 0.4cm, clip=true,width=1\textwidth]{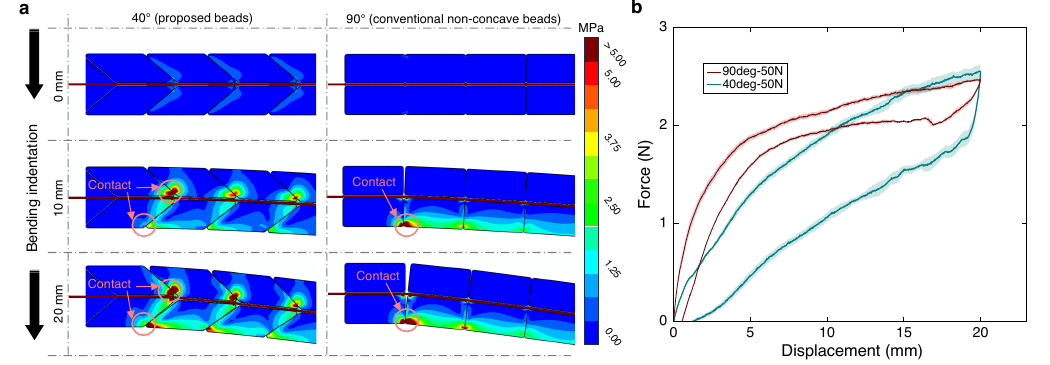}
  \caption{\textbf{Comparison of mechanical behaviors of beams with the proposed conical and conventional non-concave interfaces.} Here, we use 40$^\circ$ beads to represent our proposed method, while using 90$^\circ$ beads to exemplify conventional non-concave beads. \textbf{a}, The displacement and stress distribution over 20 mm indentation as calculated by FE analysis. \textbf{b}, The experimentally measured force-displacement curves of both two beams. The strings are pretensioned at 50 N.}
  \label{SMfig:40deg-90deg}
\end{figure*}

\begin{figure*}[ht]
  \centering
  \includegraphics[trim=0in 0mm 0in 1mm, clip=true,width=1\textwidth]{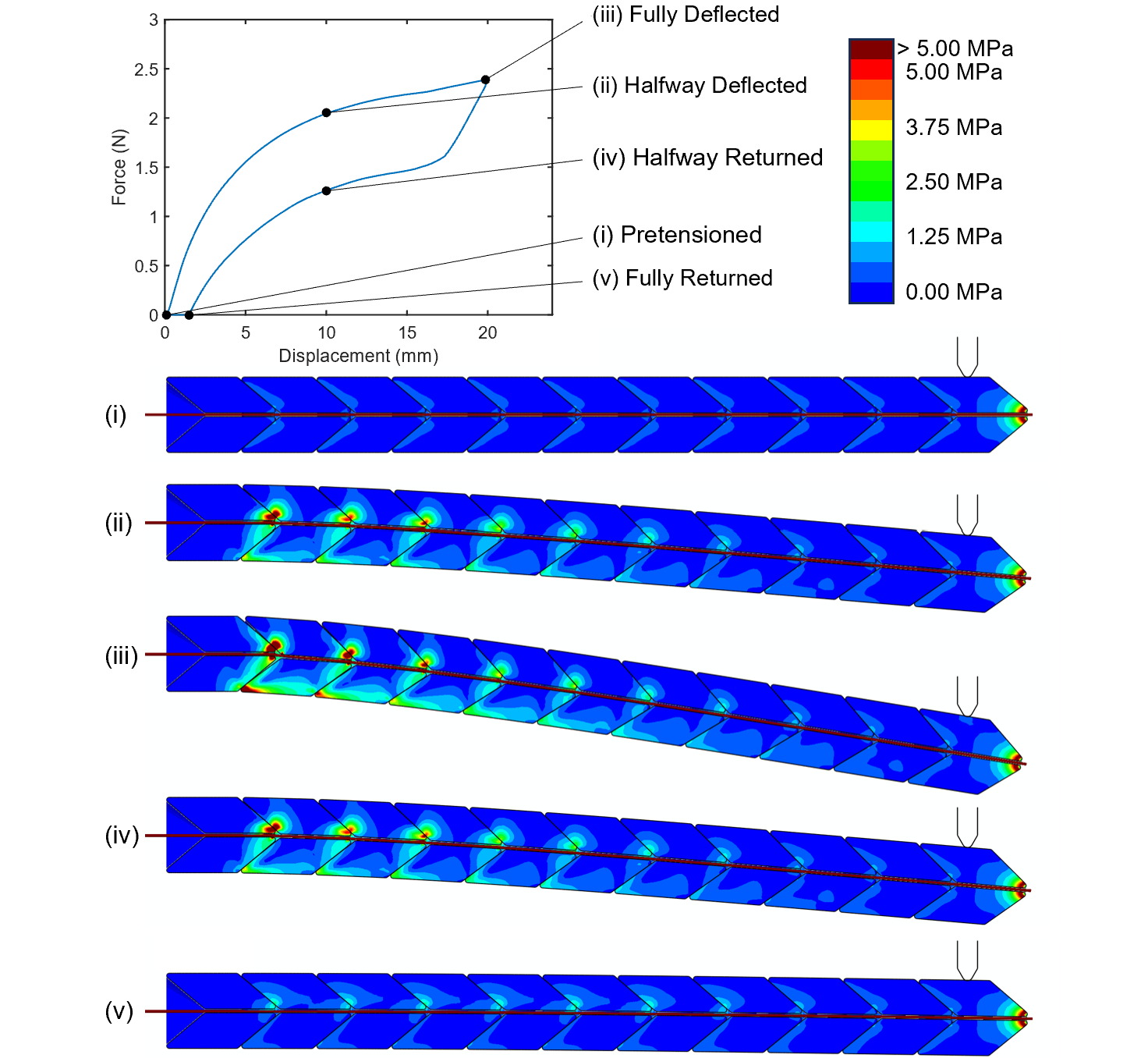}
  \caption{\textbf{Finite element simulation of a 40$^\circ$ beam under a 20 mm bending indentation.} In this model, the string was pretensioned to 50 N, the beads have an edge radius of 0.05 mm, the bead material has a Young's modulus of 0.571 GPa, and the bead-to-bead friction coefficient is 0.15. A full animation can be seen in Supplemental Movie S3.}
  \label{SMfig:fea40Bending}
\end{figure*}

\begin{figure*}[ht]
  \centering
  \includegraphics[trim=0in 0cm 0in 1mm,, clip=true,width=1\textwidth]{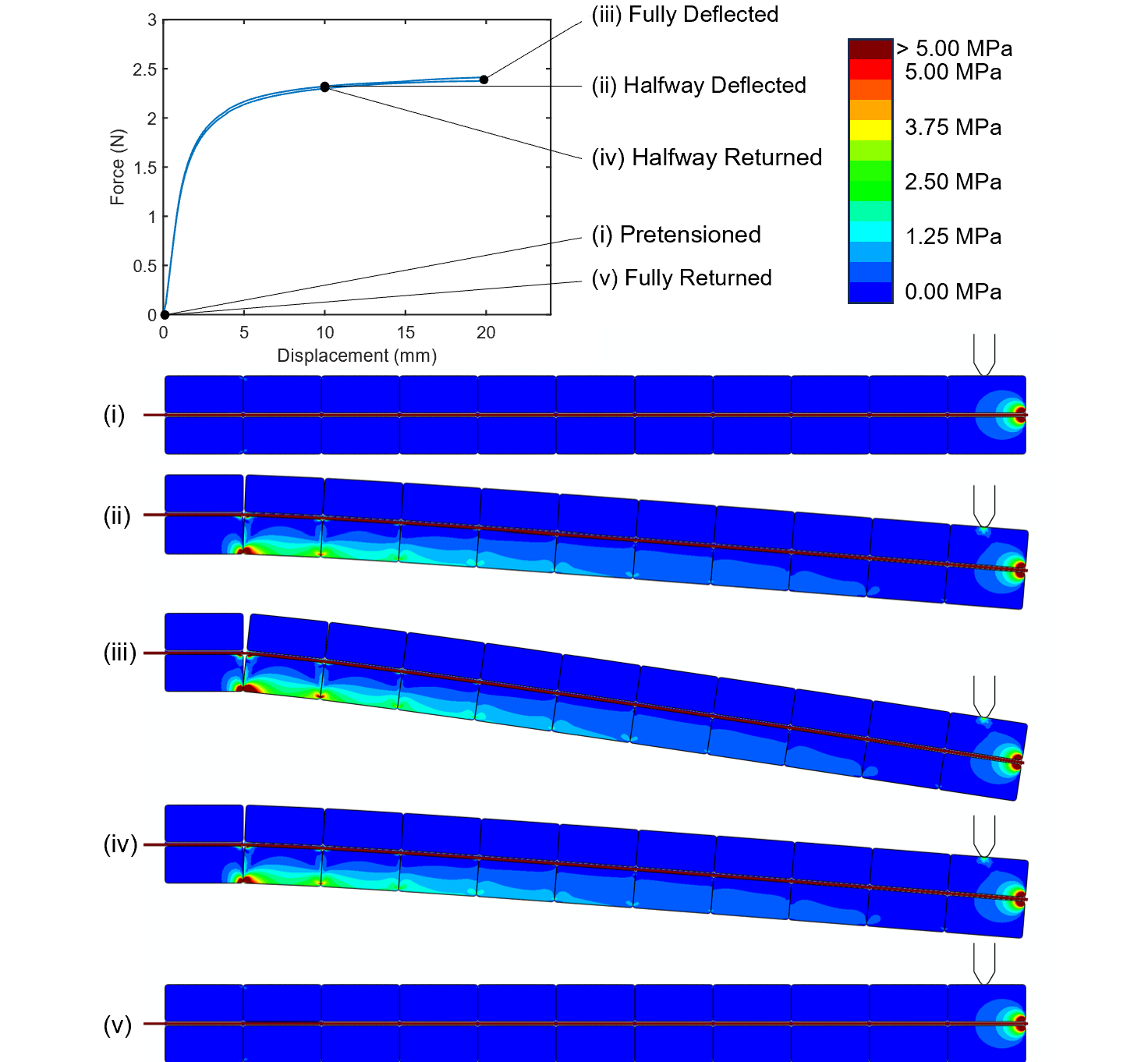}
 \caption{\textbf{Finite element simulation of a 90$^\circ$ beam under a 20 mm bending indentation.} In this model, the string was pretensioned to 50 N, the beads have an edge radius of 0.05 mm, the bead material has a Young's modulus of 0.571 GPa, and the bead-to-bead friction coefficient is 0.15. A full animation can be seen in Supplemental Movie S15.}
  \label{SMfig:fea90Bending}
\end{figure*}

\begin{figure*}[ht]
  \centering
  \includegraphics[clip=true,width=1\textwidth]{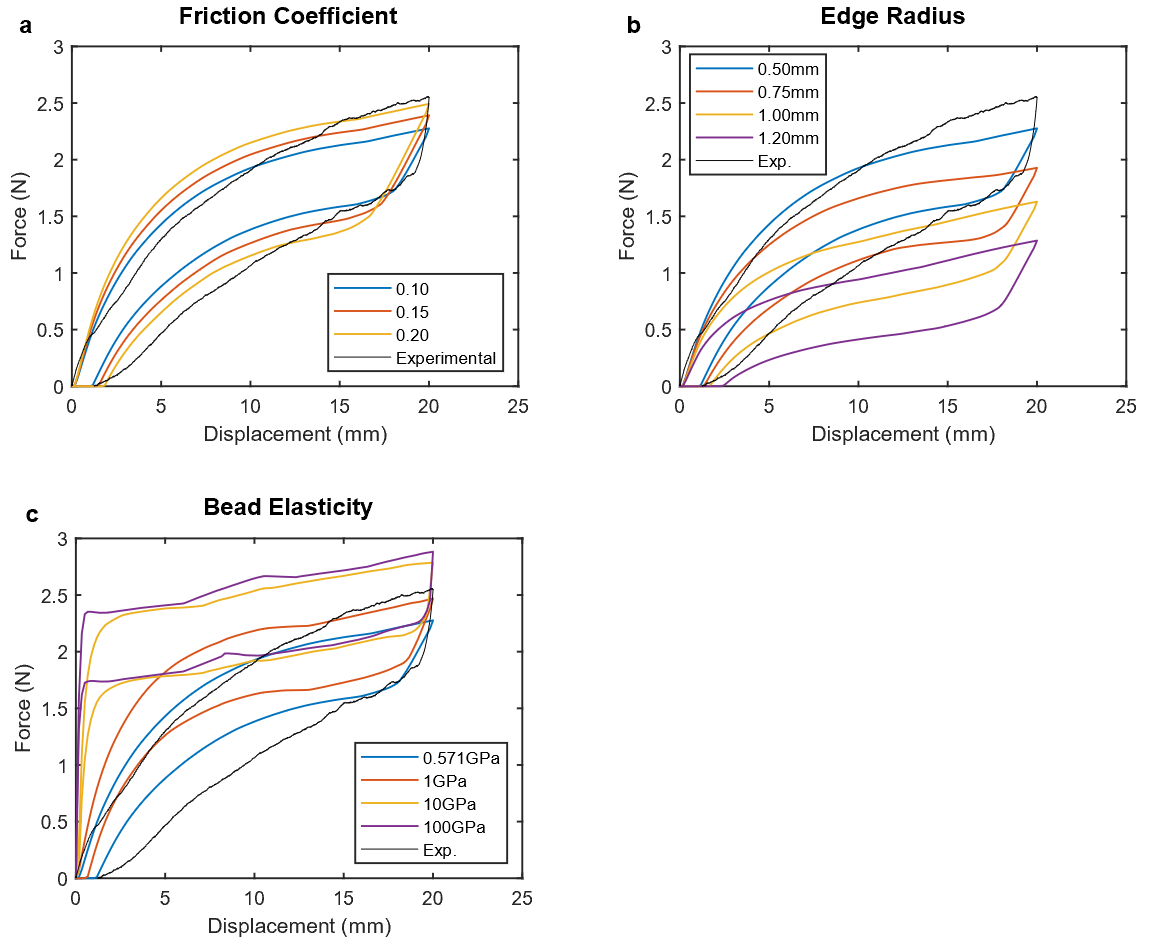}
  \caption{\textbf{Quasistatic bending FE results showing the influence of design parameters on CCPJ beams with concavo-convex beads.} Beams are subjected to 50 N of string pretension and have 40$^\circ$ cone angle beads. Unless otherwise noted, the bead edge radius is 0.5 mm, the bead Young's modulus is 0.571 GPa and the bead-to-bead friction coefficient is 0.1. Experimental data is the average of our sample data for the 50 N string tension case. \textbf{a}, Results where the bead-bead friction coefficient is varied. \textbf{b}, Results where the bead edge radius is varied. \textbf{c}, Results where the Young's modulus of the bead material is varied.}
  \label{SMfig:fea40Curves}
\end{figure*}

\begin{figure*}[ht]
  \centering
  \includegraphics[clip=true,width=1\textwidth]{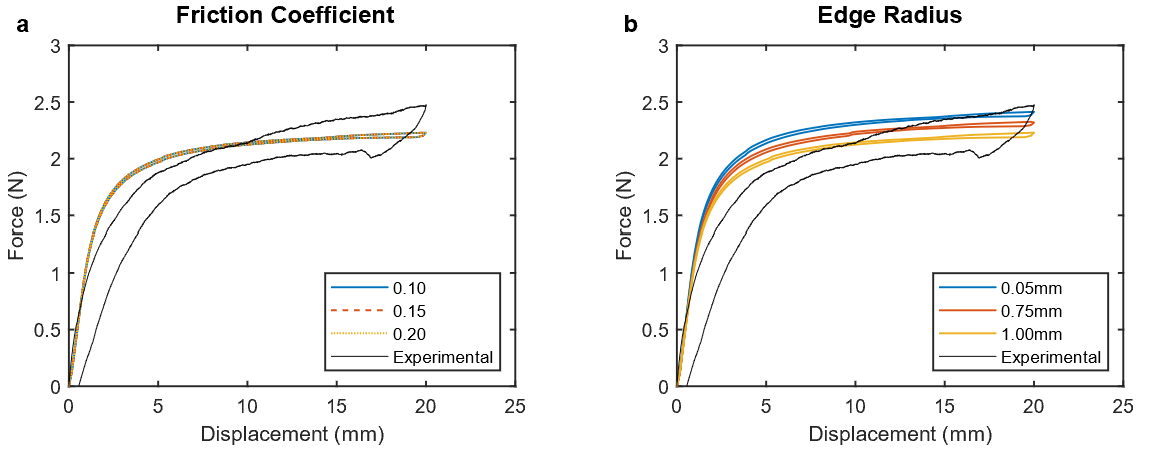}
  \caption{\textbf{Quasistatic bending FE results showing the influence of design parameters on CCPJ beams with traditional non-concave beads.} Beams are subjected to 50 N of string pretension and have 90$^\circ$ cone angle beads. Unless otherwise noted, the bead edge radius is 0.05 mm, the bead Young's modulus is 0.571 GPa and the bead-to-bead friction coefficient is 0.1. Experimental data is the average of our sample data for the 50 N string tension case. \textbf{a}, Results where the bead-bead friction coefficient is varied. \textbf{b}, Results where the bead edge radius is varied.}
  \label{SMfig:fea90Curves}
\end{figure*}

\begin{figure*}[ht]
  \centering
  \includegraphics[clip=true,width=1\textwidth]{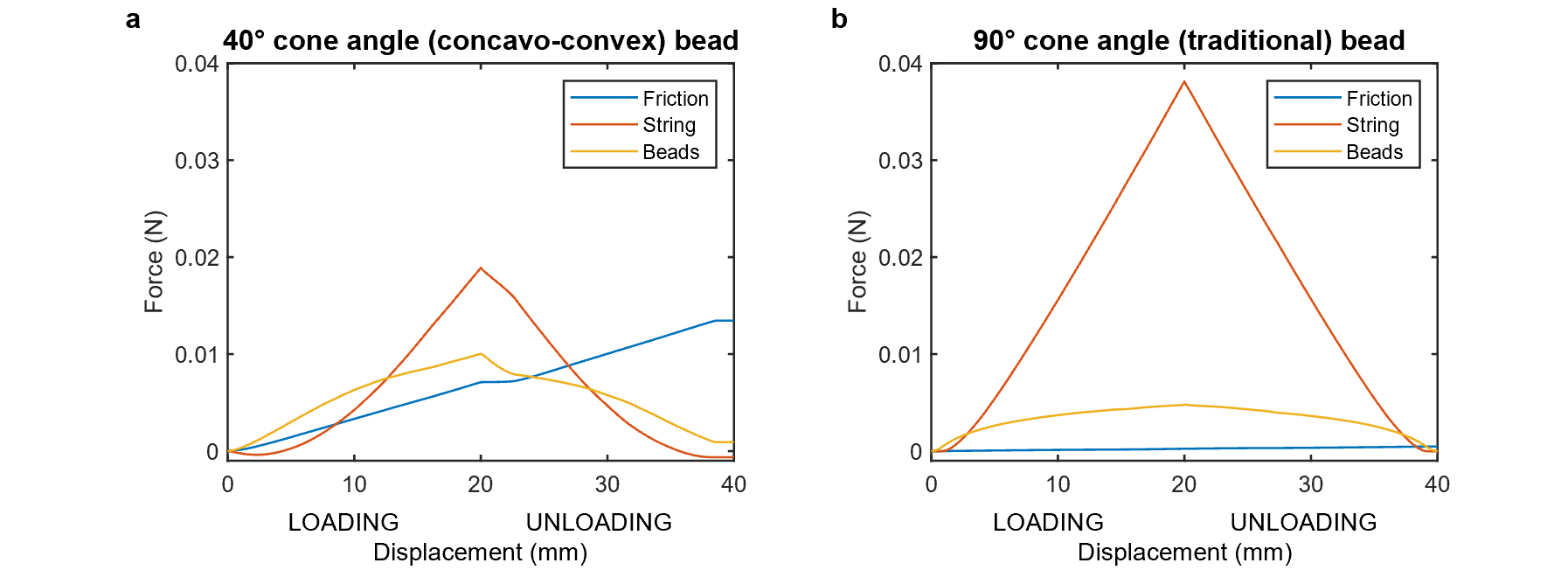}
  \caption{\textbf{Comparison of the energy dissipated by friction, the elastic energy stored in the string, and the elastic energy stored in the beads during a 20 mm loading cycle as calculated by FE methods.} \textbf{a}, Energy distribution for concavo-convex CCPJ beams with 40$^\circ$ cone angles. \textbf{b}, Energy distribution for traditional flat CCPJ beams (with 90$^\circ$ cone angles).}
  \label{SMfig:feaEnergyPlots}
\end{figure*}

\begin{figure*}[ht]
  \centering
  \includegraphics[trim=0in 1.5cm 0in 0.65cm, clip=true,width=1\textwidth]{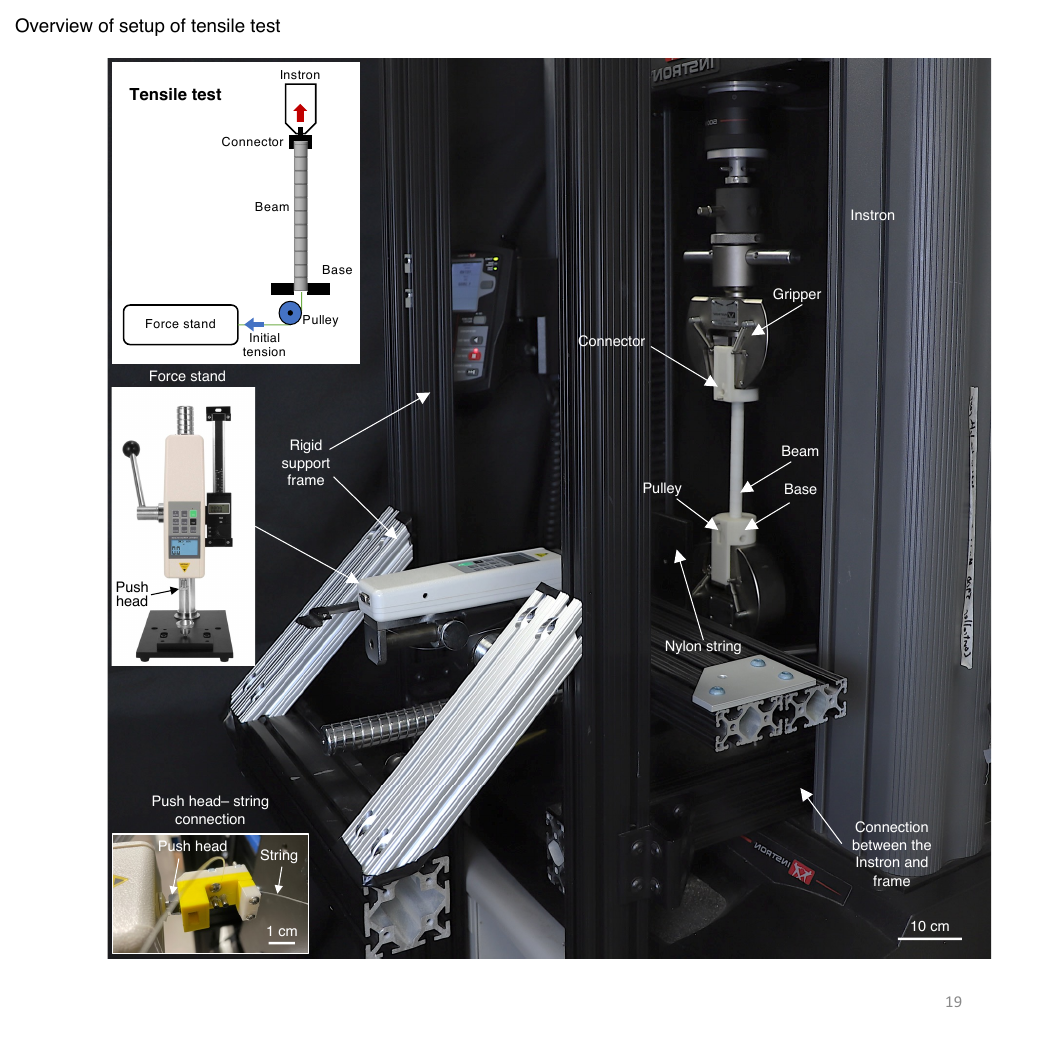}
  \caption{\textbf{Overview of the test rig.} Here we used the setup for tensile tests as an example to show the overall structure of the system. The setups for bending and compressive tests share similar structures.}
  \label{SMfig:tensileTestSetup}
\end{figure*}

\begin{figure*}[ht]
  \centering
  \includegraphics[clip=true,width=1\textwidth]{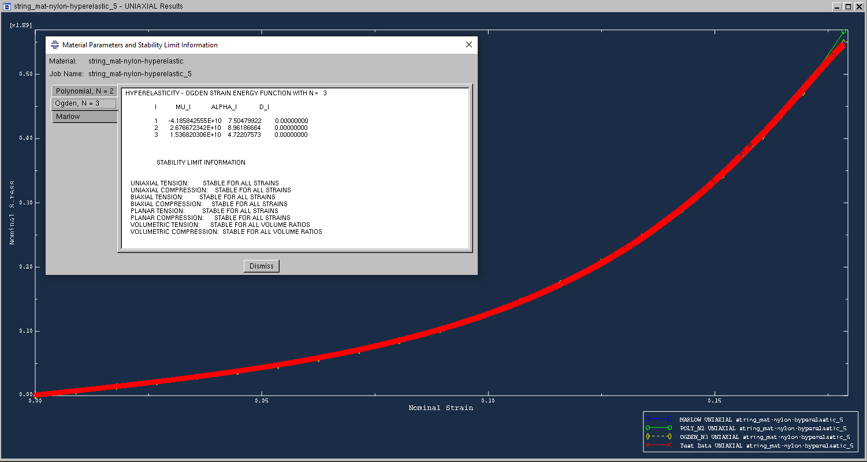}
  \caption{\textbf{Curve fitting of nylon string tensile data to Ogden model.} Hyperelastic material models fit to the test data (red) seen in Fig. \ref{SMfig:nylonTest}. Parameters for the Ogden model (yellow) in the table on the upper-left are also shown in Table \ref{table:ogden}.}
  \label{SMfig:abaqusOgdenCurveFit}
\end{figure*}

\begin{figure*}[h]
    \centering
    \includegraphics[scale=0.33]{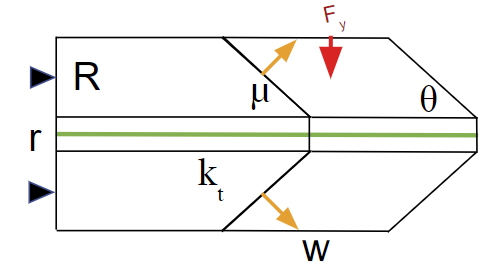}
    \includegraphics[scale=0.33]{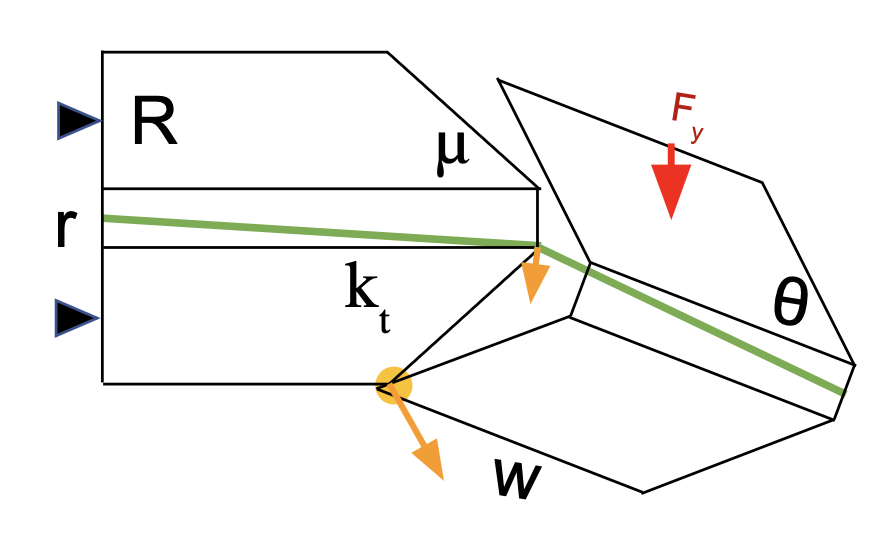}
    \includegraphics[scale=0.33]{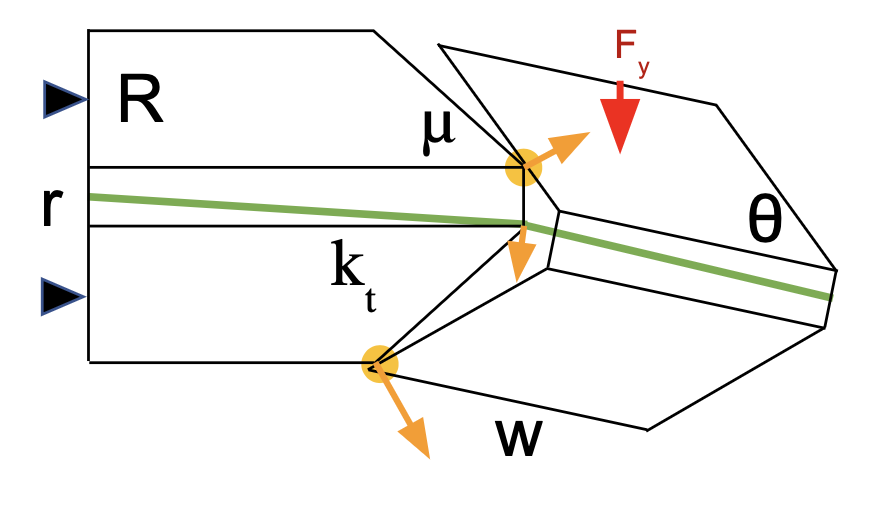}
    \caption{From left to right: surface, two-point, and one-point contact between beads for theoretical model. The forces from the bead and string are in orange. $F_y$ is the vertical loading.}
    \label{fig:schematicBead}
\end{figure*}

\begin{figure*}[h]
    \centering
    \includegraphics[trim=0in 0.5cm 0in 0.65cm, clip=true,width=0.3\textwidth]{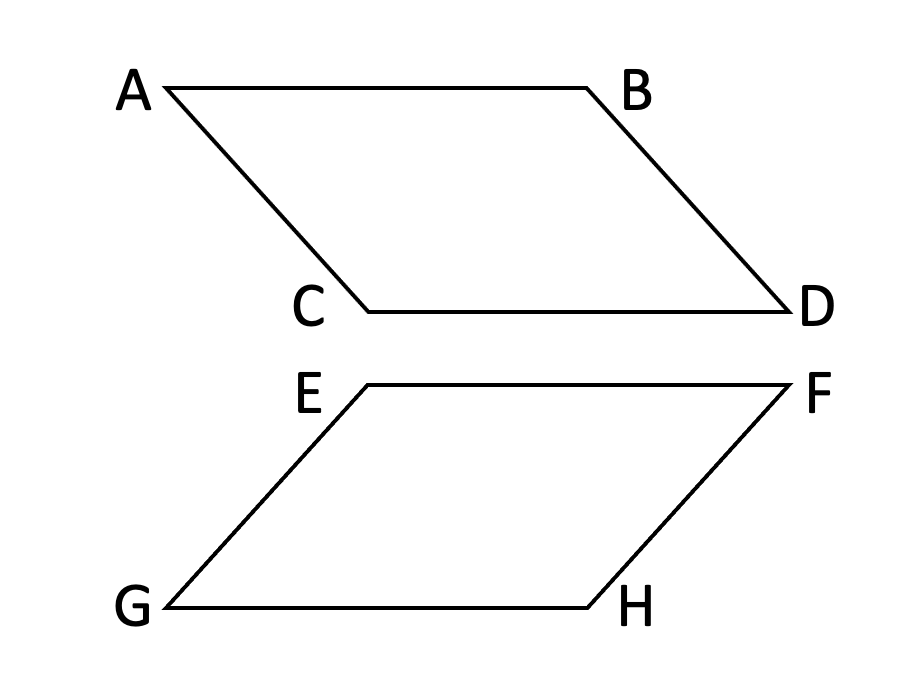}
    \caption{The bead model used by the theoretical model with points labeled.}
    \label{fig:beadModel}
\end{figure*}

\begin{figure*}[h]
    \centering
    \includegraphics[trim=4cm 7cm 4cm 5cm, clip=true, width=\textwidth]{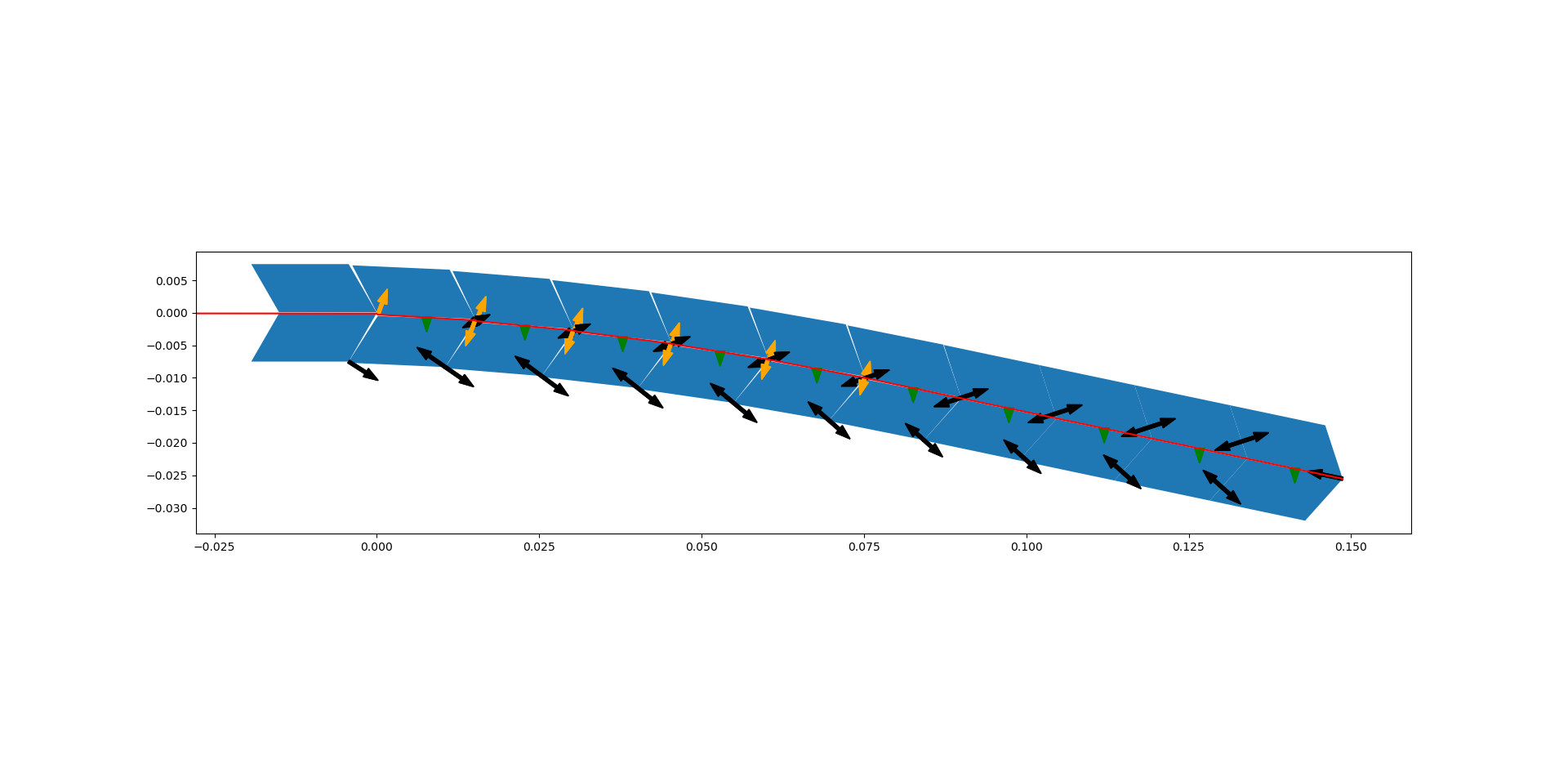}
    \caption{An example of results from the theoretical model for a beam with 40-degree beads under 50 N of string pretension. Contact forces are shown in black, string forces in orange, and gravitational forces in green.}
    \label{fig:theoreticalResult}
\end{figure*}

\begin{figure*}[h]
    \centering
    \includegraphics[width=0.7\textwidth]{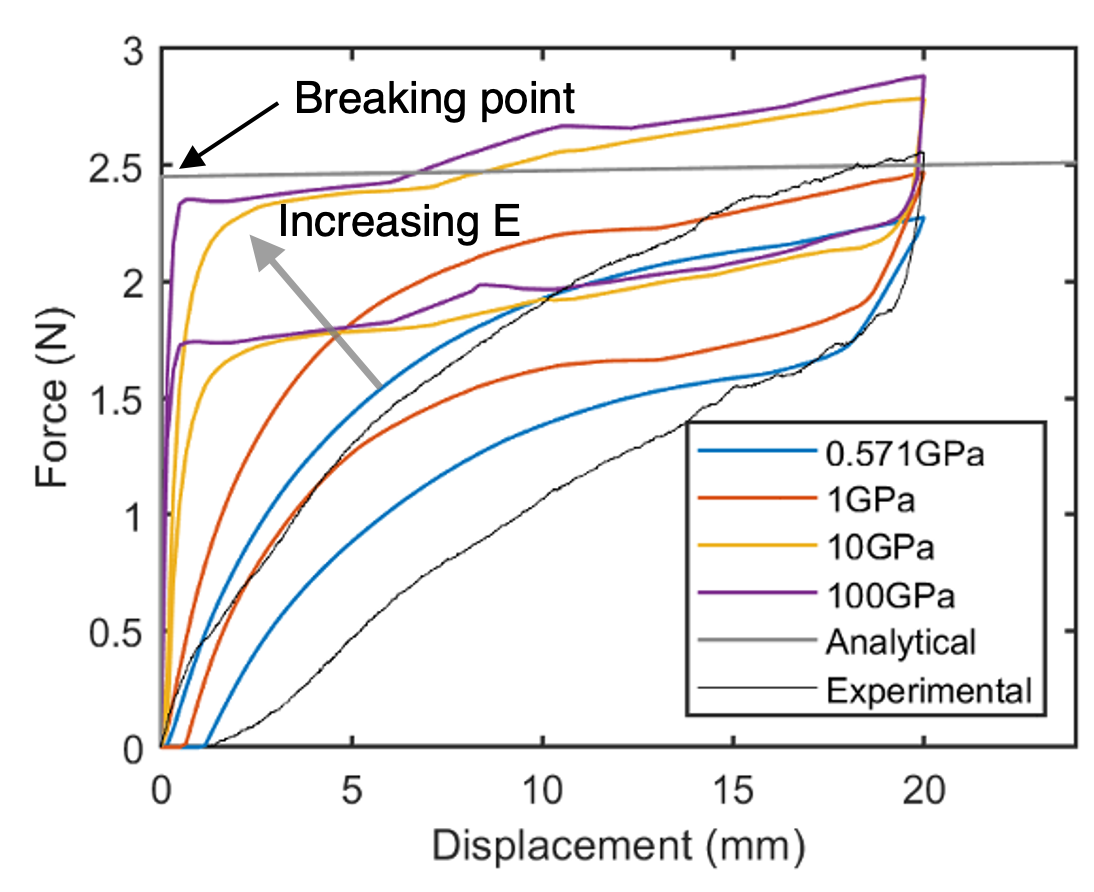}
    \caption{Comparison of experimental data, numerical results, and theoretical model. Beams are subjected to 50 N of string pretension and have 40$^\circ$ cone angle beads.}
    \label{fig:comparison}
\end{figure*}

\begin{figure*}[h]
    \centering
    \includegraphics[trim=6cm 5cm 0cm 1cm, clip=true, width=0.9\textwidth]{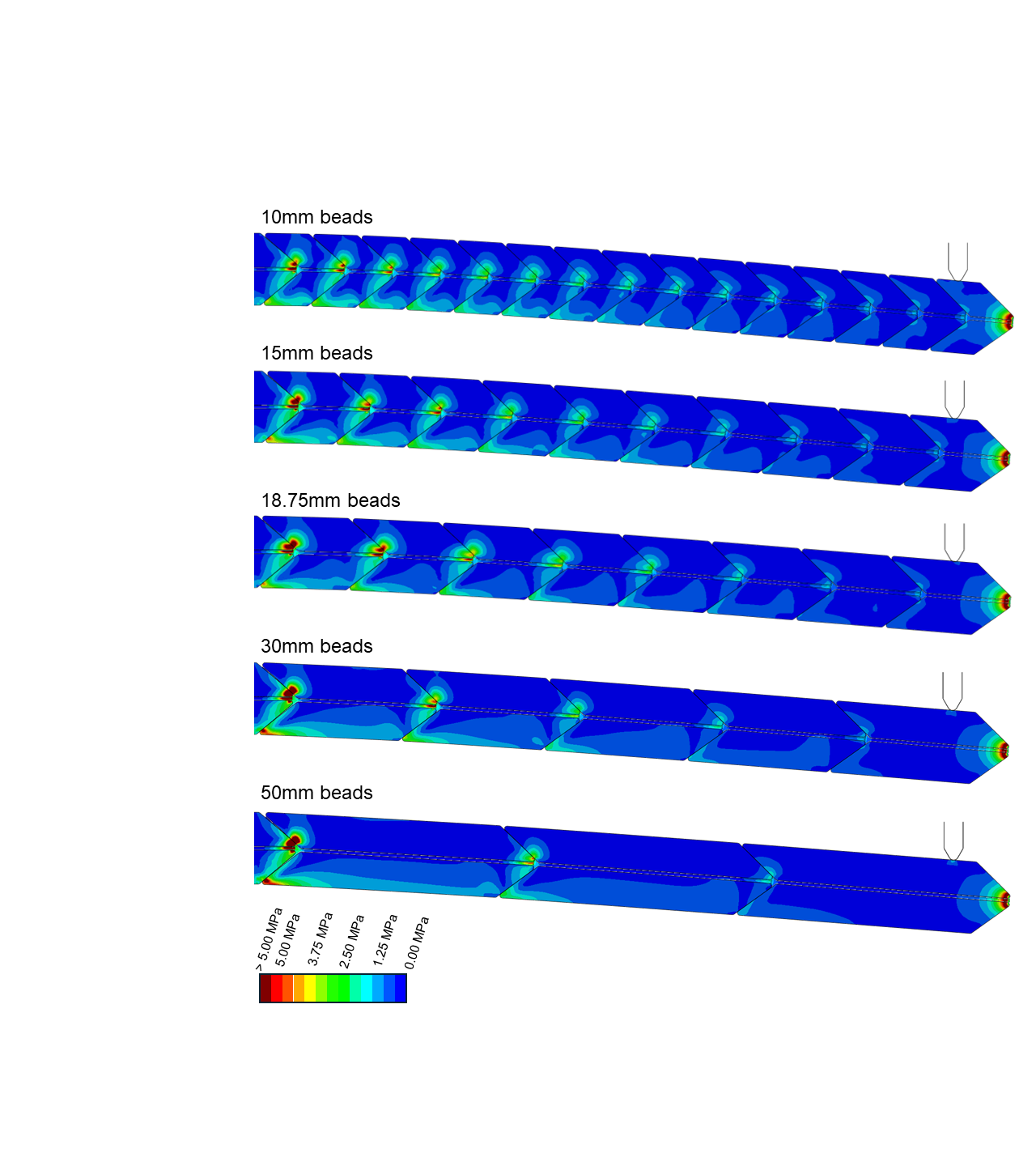}
    \caption{FE simulation of CCPJ beams with different bead lengths under 10 mm of deflection. Beams are subjected to 50 N of string pretension and have 40$^\circ$ cone angle beads. Note that the left-most bead in each beam is held fixed.}
    \label{fig:FEALength}
\end{figure*}

\begin{figure*}[h]
    \centering
    \includegraphics[trim=0cm 0.03cm 0.1cm 0cm, clip=true, width=0.6\textwidth]{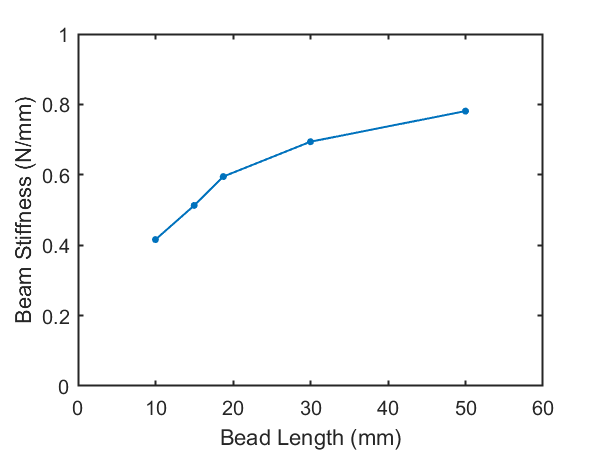}
    \caption{The effect of bead length on the stiffness of a beam under bending deflection. Beams are subjected to 50 N of string pretension and have 40$^\circ$ cone angle beads.}
    \label{fig:effectLength}
\end{figure*}

\clearpage

\subsection*{Supplementary Tables}

\begin{table}[htb]
\small
\caption{\textbf{The set of parameters of the beam for mechanical properties characterization.}}
\centering
\begin{tabular}{lllll}
\noalign{\smallskip} \hline \hline \noalign{\smallskip}
Parameters & & Unit & & Value\\
\noalign{\smallskip} \hline \hline \noalign{\smallskip}
$Bead:$\\
\noalign{\smallskip} \hline \noalign{\smallskip}
Number & &  & & 11\\
Length ($L$) & & mm & & 15.0\\
Outer diameter ($D_O$)  & & mm & & 15.0\\
Inner diameter ($D_I$)  & & mm & & 0.78\\
Cone angle ($\alpha$) & & $^{\circ}$ & & 40\\
Young's modulus  & & GPa & & 0.571\\
Poisson's ratio & & 1 & & 0.4\\
\noalign{\smallskip} \hline \noalign{\smallskip}
$Nylon\hspace{0.15cm}string:$ \\
\noalign{\smallskip} \hline \noalign{\smallskip}
Length ($x\textsubscript{0}$) & & mm & & 380.0\\
Diameter ($d$) & & mm & & 0.55\\
Young's modulus  & & GPa & & see \ref{SMtext:Ogden}\\
Poisson's ratio & & 1 & & 0.4\\
\noalign{\smallskip} \hline \hline \noalign{\smallskip}
\end{tabular}
\label{table:characterization}
\end{table}

\begin{table}[htb]
\small
\caption{\textbf{Comparison of the apparent bending modulus and loss factor between FE model and experiment for 40$^\circ$ beads.}}

\centering
\begin{tabular}{llllll}
\noalign{\smallskip} \hline \hline \noalign{\smallskip}
 & Experiment & & FE model (CoF\textsuperscript{1}) & & Discrepancy (\%)\\
\noalign{\smallskip} \hline \hline \noalign{\smallskip}
Apparent bending & 229.9 & & 216.6 (0.10) & & -5.8\\
modulus (MPa) &  & & \textbf{242.6 (0.15)} & & \textbf{+5.5}\\
 &  & & 263.0 (0.20) & & 14.4\\
\noalign{\smallskip} \hline \noalign{\smallskip}
Loss factor & 0.54 & & 0.34 (0.10) & & -37.0\\
&  & & \textbf{0.47 (0.15)} & & \textbf{-13.0}\\
&  & & 0.59 (0.20) & & +9.3\\
\noalign{\smallskip} \hline \hline \noalign{\smallskip}
\end{tabular}

\begin{tablenotes}[normal,flushleft]
  \item \scriptsize \textsuperscript{1} CoF represents the coefficient of friction between beads. The chosen value of CoF for FE simulation is 0.15 (in bold).
\end{tablenotes}

\label{table:CmparisonOfModulusandDamping}
\end{table}

\begin{table}[htb]
\small
\caption{\textbf{Parameters for Ogden hyperelastic material model of nylon string used in FE model.}}
\centering
\begin{tabular}{lllllll}
\noalign{\smallskip} \hline \hline \noalign{\smallskip}
$i$ & & $\mu_i$ & & $\alpha_i$ & & $D_i$\\
\noalign{\smallskip} \hline \hline \noalign{\smallskip}
1 & & -4.185$\times 10^{10}$ & & 7.504 & & 0\\
2 & & 2.676$\times 10^{10}$ & & 8.961 & & 0\\
3 & & 1.536$\times 10^{10}$ & & 4.722 & & 0\\
\noalign{\smallskip} \hline \hline \noalign{\smallskip}
\end{tabular}
\label{table:ogden}
\end{table}

\begin{landscape}
\begin{table}[htb]
\footnotesize
\centering
\caption{\textbf{Taxonomy of tendon-driven particle jamming mechanisms\textsuperscript{1}.}}
\centering
\begin{tabular}{lllllll}
\noalign{\smallskip} \hline \hline \noalign{\smallskip}
& Self-deployment\textsuperscript{2} & Precise alignment\textsuperscript{3} & 2D metamaterials\textsuperscript{4} & 3D metamaterials\textsuperscript{5} & Tunable stiffness & Tunable damping\\
\noalign{\smallskip} \hline \hline \noalign{\smallskip}
\rowcolor{lightgray}

Our work & \textcolor{green}{\Checkmark} & \textcolor{green}{\Checkmark} & \textcolor{green}{\Checkmark} & \textcolor{green}{\Checkmark} & \textcolor{green}{\Checkmark} & \textcolor{green}{\Checkmark}\\

Ref. \cite{jiang2019chain} & \textcolor{green}{\Checkmark} & \textcolor{red}{\XSolidBrush} & \textcolor{red}{\XSolidBrush}  & \textcolor{red}{\XSolidBrush}  & \textcolor{green}{\Checkmark} & \textcolor{red}{\XSolidBrush}\\

Ref. \cite{lihua2019programmable} & \textcolor{red}{\XSolidBrush} & \textcolor{red}{\XSolidBrush} & \textcolor{green}{\Checkmark} & \textcolor{green}{\Checkmark} & \textcolor{red}{\XSolidBrush} & \textcolor{red}{\XSolidBrush}\\

Ref. \cite{yang2024hierarchical}  & \textcolor{green}{\Checkmark} & \textcolor{red}{\XSolidBrush}  & \textcolor{green}{\Checkmark} & \textcolor{yellow}{\Checkmark}\textsuperscript{*} & \textcolor{green}{\Checkmark} & \textcolor{red}{\XSolidBrush}\\

Ref. \cite{Kenjiro2020radial} &  \textcolor{red}{\XSolidBrush} & \textcolor{red}{\XSolidBrush} & \textcolor{red}{\XSolidBrush} & \textcolor{red}{\XSolidBrush} & \textcolor{green}{\Checkmark} & \textcolor{red}{\XSolidBrush}\\
\noalign{\smallskip} \hline \hline \noalign{\smallskip}
\end{tabular}

\begin{tablenotes}[normal,flushleft]
  \footnotesize
  \item \footnotesize Note: 1: Here, we only list representative mechanisms, excluding traditional vacuum-driven particle jamming. 2: The systems can be deployed from compact states to preprogrammed states without requiring additional actuation or manual manipulations. 3: The deployed systems should have small alignment offsets. 4: 2D lattice structures. 5: 3D lattice structures. *: This work generates 2D surfaces in a 3D space. 
\end{tablenotes}
\label{table:compare}
\end{table}
\end{landscape}

\begin{table}[h]
    \centering
     \caption{\textbf{System parameters of bead model.}}
    \begin{tabular}{llll}
        \hline
        \textbf{Symbol} & \textbf{Value} & \textbf{unit} & \textbf{Description} \\ \hline
        $R$ & 7.5 2 & mm & Outer radius \\ \hline
        $r$ & 0.1 & mm & Inner radius \\ \hline
        $w$ & 1.5 & cm & Bead length \\ \hline
        $\theta$ & $4 \pi/18$ & rad & Bead angle (here 40 deg) \\ \hline
        $\mu$ & 0.1 & - & Friction coefficient \\ \hline
        $\rho$ & 1000 & $kg/m^3$ & Bead density \\ \hline
        $n_{beads}$ & 10 & - & Number of beads (not counting base) \\ \hline
        $T_s$ & 50 & N & Initial tension in string \\ \hline
    \end{tabular}
    \label{tab:parametersOfBeads}
\end{table}

\clearpage

\subsection*{Description of Supplementary Movies}
\bigskip
\noindent {File name: Supplementary Movie S1 \\ Description: Self-deployment and mechanical properties manipulation of a unit beam.} 

\bigskip
\noindent {File name: Supplementary Movie S2 \\ Description: Bending test of a beam composed of eleven resin beads with 40$^{\circ}$ cone angle. The initial tension on the string is 50 N.} 

\bigskip
\noindent {File name: Supplementary Movie S3 \\ Description: Cross-section side view of the von Mises stress plot animation of a beam constructed of eleven 40$^{\circ}$ beads in a finite element model bending test. The contracting tension is 50 N.}

\bigskip
\noindent {File name: Supplementary Movie S4 \\ Description: Tensile test of a beam composed of eleven resin beads with 40$^{\circ}$ cone angle. The contracting tension is 50 N.} 

\bigskip
\noindent {File name: Supplementary Movie S5 \\ Description: Compressive test of a beam composed of eleven resin beads with 40$^{\circ}$ cone angle. The contracting tension is 50 N.}

\bigskip
\noindent {File name: Supplementary Movie S6 \\ Description: 
Cable-driven self-deployment tests of resin beams against gravity.}

\bigskip
\noindent {File name: Supplementary Movie S7 \\ Description: Compressive test of unit cells (both bending- and stretching-dominated) composed of beams with 40$^{\circ}$ resin beads. The contracting tension is 40 N.}

\bigskip
\noindent {File name: Supplementary Movie S8 \\ Description: Self-deployment/collapse of a beam with 40$^{\circ}$ resin beads, activated by a motor-driven cable.}

\bigskip
\noindent {File name: Supplementary Movie S9 \\ Description: Self-deployment/collapse of a beam with 90$^{\circ}$ plywood beads, activated by an electrically driven artificial muscle, super-coiled polymer actuator.}

\bigskip
\noindent {File name: Supplementary Movie S10 \\ Description: Self-deployment/collapse of a cubic lattice composed of with 90$^{\circ}$ plywood beads and electrically driven artificial muscles---super-coiled polymer actuators.}

\bigskip
\noindent {File name: Supplementary Movie S11 \\ Description: Self-deployment, loading-bearing, and self-collapse of a metamaterial composed of an arrangement of $2 \times 2 \times 2$ cubic unit cell.}

\bigskip
\noindent {File name: Supplementary Movie S12 \\ Description: Self-deployment, mechanical property tuning, and self-collapse of a metamaterial composed of an arrangement of $2 \times 2 \times 2$ cubic unit cell.}

\bigskip
\noindent {File name: Supplementary Movie S13 \\ Description: View of the von Mises stress plot animation of a beam constructed of eleven 40$^{\circ}$ beads in a finite element model bending test. The contracting tension is 50 N.}

\bigskip
\noindent {File name: Supplementary Movie S14 \\ Description: View of the von Mises stress plot animation of a beam constructed of eleven 90$^{\circ}$ beads in a finite element model bending test. The contracting tension is 50 N.}

\bigskip
\noindent {File name: Supplementary Movie S15 \\ Description: Cross-section side view of the von Mises stress plot animation of a beam constructed of eleven 90$^{\circ}$ beads in a finite element model bending test. The contracting tension is 50 N.}

\bigskip
\noindent {File name: Supplementary Movie S16 \\ Description: Bending test of a beam composed of eleven resin beads with 90$^{\circ}$ cone angle. The initial tension on the string is 50 N.}

\clearpage

\end{document}